%% file: Main.tex
\documentclass[acmlarge,screen]{acmart}
\AtBeginDocument{%
  }

\usepackage{subcaption}
\usepackage{wasysym}
\usepackage{makecell}
\usepackage{enumitem}
\usepackage{soul}
\usepackage{color, xcolor}
\usepackage{pifont}
\usepackage{algorithm}
\usepackage{algpseudocode}
\usepackage[most]{tcolorbox}
\usepackage{etoolbox}
\apptocmd{\thebibliography}{\interlinepenalty=10000\relax}{}{}

\newcommand{\revision}[1]{{\color{black}{#1}}}
\newcommand{\final}[1]{{\color{black}{#1}}}

\makeatletter
  \let\originalfnsymbol\@fnsymbol
  \renewcommand{\@fnsymbol}[1]{%
    \ifcase#1\or
      \originalfnsymbol{2}%
    \or
      \originalfnsymbol{1}%
    \else
      \originalfnsymbol{#1}%
    \fi}
\makeatother
\newcommand{\workname}{\texorpdfstring{Pro\texttwosuperior{}Assist}{Pro2Assist}}

\begin{document}

\title{\workname: Continuous Step-aware Proactive Assistance with Multi-modal Egocentric Perception for Long-horizon  Procedural Tasks}

\author{Lilin Xu}
\orcid{0009-0007-5203-7496}
\authornote{Co-author equal contribution.}
\affiliation{%
  \institution{Columbia University}
  \city{New York}
  \country{United States}}
\email{lilin.xu@columbia.edu}

\author{Bufang Yang}
\orcid{0000-0003-0032-2539}
\authornotemark[1]
\affiliation{%
  \institution{The Chinese University of Hong Kong}
  \city{Hong Kong}
  \country{China}}
\email{bfyang@link.cuhk.edu.hk}

\author{Siyang Jiang}
\orcid{0000-0002-9926-6532}
\affiliation{%
  \institution{The Chinese University of Hong Kong}
  \city{Hong Kong}
  \country{China}}
\email{syjiang@ie.cuhk.edu.hk}

\author{Kaiwei Liu}
\orcid{0009-0002-4108-0898}
\affiliation{%
  \institution{The Chinese University of Hong Kong}
  \city{Hong Kong}
  \country{China}}
\email{1155189693@link.cuhk.edu.hk}

\author{Kaiyuan Hou}
\orcid{0000-0002-6495-401X}
\affiliation{%
  \institution{Columbia University}
  \city{New York}
  \country{United States}}
\email{kh3119@columbia.edu}

\author{Yuang Fan}
\orcid{0000-0002-0976-9292}
\affiliation{%
  \institution{Columbia University}
  \city{New York}
  \country{United States}}
\email{yf2676@columbia.edu}

\author{Hongkai Chen}
\orcid{0000-0001-7206-6584}
\affiliation{%
  \institution{The Chinese University of Hong Kong}
  \city{Hong Kong}
  \country{China}}
\email{hkchen@ie.cuhk.edu.hk}

\author{Zhenyu Yan}
\orcid{0000-0002-4433-5211}
\affiliation{%
  \institution{The Chinese University of Hong Kong}
  \city{Hong Kong}
  \country{China}}
\email{zyyan@ie.cuhk.edu.hk}

\author{Xiaofan Jiang}
\orcid{0000-0002-6480-0299}
\authornote{Corresponding author.}
\affiliation{%
  \institution{Columbia University}
  \city{New York}
  \country{United States}}
\email{jiang@ee.columbia.edu}

\authorsaddresses{%
Authors' Contact Information:
\href{https://orcid.org/0009-0007-5203-7496}{Lilin Xu}, Columbia University, New York, United States, lilin.xu@columbia.edu;
\href{https://orcid.org/0000-0003-0032-2539}{Bufang Yang}, The Chinese University of Hong Kong, Hong Kong, China, bfyang@link.cuhk.edu.hk;
\href{https://orcid.org/0000-0002-9926-6532}{Siyang Jiang}, The Chinese University of Hong Kong, Hong Kong, China, syjiang@ie.cuhk.edu.hk;
\href{https://orcid.org/0009-0002-4108-0898}{Kaiwei Liu}, The Chinese University of Hong Kong, Hong Kong, China, 1155189693@link.cuhk.edu.hk;
\href{https://orcid.org/0000-0002-6495-401X}{Kaiyuan Hou}, Columbia University, New York, United States, kh3119@columbia.edu;
\href{https://orcid.org/0000-0002-0976-9292}{Yuang Fan}, Columbia University, New York, United States, yf2676@columbia.edu;
\href{https://orcid.org/0000-0001-7206-6584}{Hongkai Chen}, The Chinese University of Hong Kong, Hong Kong, China, hkchen@ie.cuhk.edu.hk;
\href{https://orcid.org/0000-0002-4433-5211}{Zhenyu Yan}, The Chinese University of Hong Kong, Hong Kong, China, zyyan@ie.cuhk.edu.hk;
\href{https://orcid.org/0000-0002-6480-0299}{Xiaofan Jiang} (corresponding author), Columbia University, New York, United States, jiang@ee.columbia.edu.}
\renewcommand{\shortauthors}{Xu et al.}

\renewcommand{\shorttitle}{\workname: Continuous Step-aware Proactive Assistance with Multimodel Egocentric Perception}

\input{section/Abstract}

\setcopyright{cc}
\setcctype{by}
\acmJournal{IMWUT}
\acmYear{2026} \acmVolume{10} \acmNumber{3} \acmArticle{178}
\acmMonth{9} \acmDOI{10.1145/3832022}



\settopmatter{printfolios=true}
\maketitle
\newcommand{\cmt}[1]{\textcolor{black}{#1}}
\newcommand{\xl}[1]{{\color{magenta}{#1}}}

\input{section/Introduction.tex}

\input{section/Related_work.tex}

\input{section/Motivation.tex}

\input{section/System_design.tex}

\input{section/Evaluation.tex}
\input{section/Discussion.tex}
\input{section/Conclusion.tex}

\section*{Acknowledgments}
This paper was supported in part by a Columbia Engineering Presidential Fellowship. This work was also partially supported by the Research Grants Council of Hong Kong under grants STG1/E-403/24-N and 14212323.

\section*{Use of Generative AI}
\final{Generative AI tools (OpenAI ChatGPT and Anthropic Claude) were used to improve the quality of writing, including style, phrasing, and grammar polishing of author-written text. Typical prompts included variations of ``improve the writing of the following sentences to make it concise and clear without changing its meaning''. All AI-suggested edits were reviewed, verified, and revised by the authors.
}

\bibliographystyle{ACM-Reference-Format}
\bibliography{ref}

\newpage
\appendix
\input{section/Appendix}

\end{document}

%% file: section/Abstract.tex
\begin{abstract}
Procedural tasks with multiple ordered steps are ubiquitous in daily life. 
Recent advances in multimodal large language models (MLLMs) have enabled personal assistants that support daily activities.
However, existing systems primarily provide reactive guidance triggered by user queries, or limited proactive assistance for isolated short-term events rather than long-horizon procedural tasks.
In this work, we introduce \workname, a step-aware proactive assistant that \revision{continuously tracks fine-grained task progress and reasons over the user's evolving state to provide timely assistance throughout tasks.}
\workname~leverages multimodal data from augmented reality (AR) glasses to achieve \revision{motion-based perception}.
It then extracts step-oriented procedural context from multi-scale temporal dynamics and task-specific expert knowledge. 
Based on both sensory input and procedural context, \workname~performs continuous reasoning to infer user needs and display timely assistance on AR glasses.
We evaluate \workname~using a dataset curated from public sources and a real-world dataset collected on our testbed with AR glasses.
Extensive evaluations show that \workname~outperforms the best-performing baselines by over 21\% in procedural action understanding accuracy, and it achieves up to 2.29$\times$ the proactive timing accuracy of baselines.
A user study with 20 participants further shows that 90\% find \workname~useful, indicating its effectiveness for real-world procedural assistance.

\end{abstract}

%% file: section/Introduction.tex
\section{INTRODUCTION}
Procedural tasks are ubiquitous in daily life and play a crucial role in many routine human activities, spanning from cooking to assembling everyday items~\cite{arakawa2024prism,lee2024error}. 
These tasks typically involve multiple steps that need to be executed in precise order, which can be challenging when the procedure is complex or unfamiliar to the user.
Although there are usually instruction manuals and online tutorials available, they require users to repeatedly shift attention between physical actions and external references, leading to cognitive interruptions and increased mental load~\cite{raouf1980effect, tang2003comparative}.
With the rapid advancement of LLMs and MLLMs~\cite{singh2025openai,Qwen3-VL,gemini}, intelligent personal assistants have been developed to handle users' questions by providing relevant task instructions~\cite{huang2025vinci,arakawa2024prism}, thereby reducing the need for manual instruction lookup during procedural tasks.

However, most of these procedural assistants are reactive, requiring explicit user queries that interrupt ongoing actions and undermine seamless task guidance.
Compared to reactive assistants, recent research has proposed proactive systems~\cite{yang2025proagent,yang2025contextagent,liu2024chainstream} that aim to further reduce users’ physical and mental workload by inferring when and what assistance to provide without waiting for explicit queries.
These systems can recognize short-term events and proactively assist with them, such as detecting a user viewing products and offering price comparisons.
However, most existing proactive systems provide one-shot\footnote{Throughout this paper, ``one-shot'' refers to providing isolated assistance for the overall event based on holistic scene understanding} assistance at the level of an isolated event based on holistic scene understanding, rather than continuous, step-by-step guidance. 
As a result, they are less suitable for long-horizon procedural tasks with multiple steps, where user needs evolve over time and are strongly correlated with task progress, as illustrated on the right side of Figure~\ref{fig:tesear}.
\revision{Although recent works~\cite{arakawa2024prismobserver,arakawa2025scaling,li2025satori} explore proactive interventions for procedural tasks, they rely on discrete trigger events and primarily deliver content about the next step.}
\revision{These gaps highlight the need for \textit{continuous, step-aware} assistants that understand the user's ongoing actions and deliver timely guidance grounded in the user's actual state throughout long-horizon procedural tasks.}

\revision{To address this gap, we aim to develop a procedural assistive system that continuously tracks fine-grained task progress, including the current procedural step and within-step execution state (e.g., just started or about to finish), and reasons over the user's evolving state to help users perform tasks smoothly.
However, developing such a system introduces several unique challenges.}
\textit{First,} procedural tasks involve continuous interactions between the user and the physical environment, where the user's attention and state are crucial for identifying assistive moments.
However, existing works~\cite{chen2024videollm,wu2024videollm} mainly rely on egocentric vision and often overlook implicit attention cues from head and hand motion, which leads to limited intent and attention understanding, weakening the timeliness of assistance.
\textit{Second,} \revision{continuously tracking fine-grained task} progress requires capturing temporal dynamics, including short-term hand manipulations and long-term historical context, together with procedural knowledge, rather than relying on \revision{isolated single-moment observations}.
However, prior work~\cite{chen2024videollm} mainly relies on vision-only dense frame sequences to capture temporal context without explicitly modeling user intent or procedural knowledge, posing challenges for correctly interpreting fine-grained task progress.
\textit{Third,} while existing VLMs~\cite{ye2024mm,zhou2025egotextvqa,vinod2025egovlm} are developed for general scene and action understanding, they exhibit insufficient capability in \revision{understanding users' ongoing actions in long-horizon and temporally dependent procedural tasks.}
\revision{This makes it difficult to continuously reason over the user's evolving state, limiting the ability to deliver assistance grounded in the user's actual state.}

\input{insert_figures/teaser}

In this paper, we introduce \workname, a continuous, step-aware \underline{Pro}active \underline{Assist}ant for long-horizon \underline{Pro}cedural tasks that integrates multimodal egocentric perception and LLM reasoning to deliver timely assistance through AR glasses, as shown in Figure~\ref{fig:tesear}'s left side.
\workname~first introduces an \revision{motion-based} perception mechanism based on multimodal egocentric data from AR glasses.
It integrates head-motion-aware sampling with key moment selection based on optical flow estimation over visual data to identify moments with high potential for assistive needs \revision{from continuous observations.}
Next, \workname~performs step-oriented procedural context extraction by incorporating task-specific expert knowledge together with multi-scale temporal context, including short-term hand motion cues and long-term task progress.
Finally, combining the extracted procedural context with sensory context, \workname~performs step-aware proactive reasoning to infer users’ current states and needs, enabling \revision{continuous} assistance aligned with fine-grained task progress.
To ensure reliable and non-intrusive assistance \revision{under continuous reasoning}, \workname~also introduces a step-aware consistency checking mechanism that utilizes historical predictions to mitigate single-moment errors and improve response timing.

We implement \workname~on a real-world testbed with AR glasses and a back-end server.
We conduct extensive evaluations on both a dataset curated from three public datasets and a real-world dataset collected on our testbed.
Results show that \workname~achieves effective proactive step-aware assistance, significantly outperforming state-of-the-art baselines.
Compared with the best-performing baselines, \workname~improves step identification accuracy by 25.2\%, execution status identification accuracy by 21.6\%, and proactive accuracy by 15.1\%, and it achieves up to 2.29$\times$ the proactive timing accuracy of the baselines.
The results also show that \workname~is robust across different VLM scales and system settings.
In addition, a user study with 20 participants indicates that 90\% found \workname~useful, with particularly strong agreement among users unfamiliar with the tasks, highlighting its practical effectiveness.
We summarize the contributions of this paper as follows.
\begin{itemize}[leftmargin=*]
\item We present \workname, an end-to-end assistive system that delivers \revision{continuous}, step-aware assistance throughout procedural tasks by \revision{observing and reasoning over} multimodal egocentric data from AR smart glasses.

\item We propose a \revision{motion-based} perception mechanism and a step-oriented procedural context extraction approach \revision{for efficient continuous perception and fine-grained task progress tracking.}
\workname~first utilizes attention cues \final{by combining head motion signals with visual motion cues} to identify moments that are likely to require proactive assistance.
It then effectively extracts procedural context by integrating multi-scale temporal context with task-specific expert knowledge.

\item 
We develop a step-aware proactive reasoner \final{that reasons over the user’s evolving task state rather than isolated observations,} with a consistency checking mechanism \final{that leverages temporal consistency in continuous reasoning}, enabling timely and non-intrusive assistance aligned with fine-grained task progress.

\item 
We implement \workname~on a real-world testbed with AR glasses and a back-end server.
Extensive evaluations on both a curated dataset and a real-world dataset demonstrate that \workname~significantly outperforms state-of-the-art baselines in procedural action understanding and proactive assistance performance, highlighting its effectiveness in delivering timely and step-aware proactive assistance.
A user study shows 90\% of participants find \workname~useful, further indicating its practical effectiveness.

\end{itemize}

%% file: insert_figures/teaser.tex
\begin{figure}[t]
    \centering
    \includegraphics[width=\textwidth]{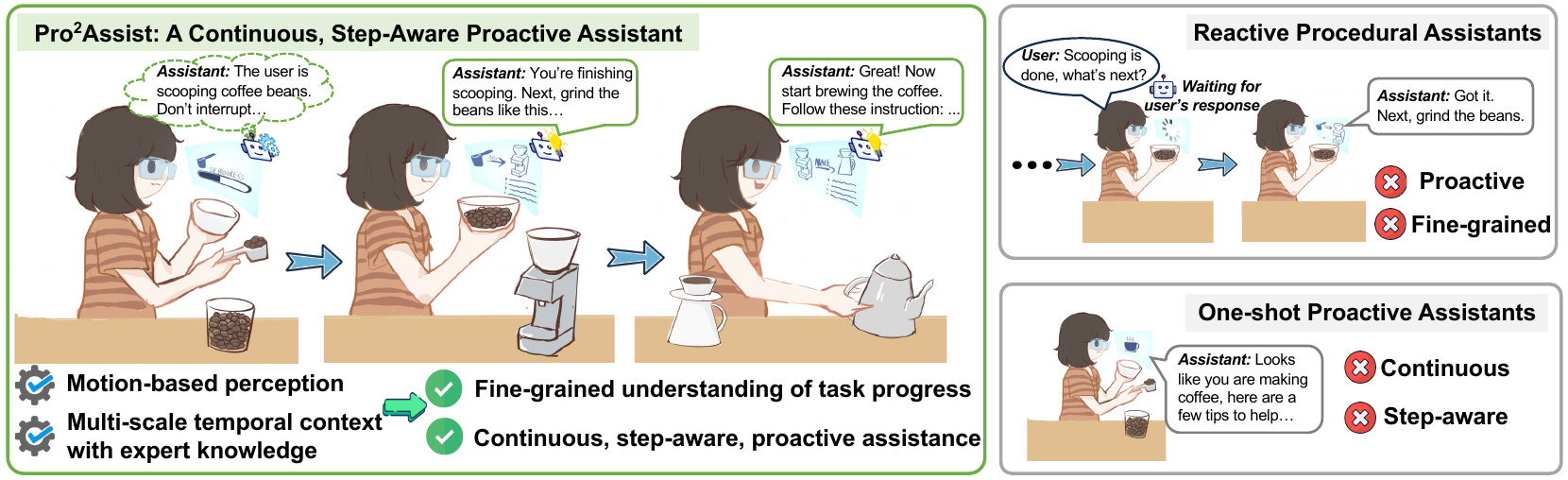}
    \caption{Application scenario of \workname. \workname~provides continuous, step-aware proactive assistance by reasoning over the user’s evolving state in the task workflow (left), while reactive procedural assistants rely on explicit user requests and one-shot proactive assistants provide isolated assistance for the overall event (right).} 
    \label{fig:tesear}
\end{figure}

%% file: section/Related_work.tex
\section{RELATED WORK}
\label{sec:related_work}

\subsection{Egocentric Smart Assistants}
\revision{Egocentric perception has been widely adopted for applications such as object recognition~\cite{akiva2023self,wu2023label}, action recognition~\cite{kukleva2024x}, and error detection~\cite{chan2024detecting,lee2024error}.
Beyond these task-specific applications, recent studies have developed egocentric smart assistants that leverage large language and vision models to support everyday human activities~\cite{xu2024can,huang2025vinci,yang2025socialmind}.}
MM-Ego~\cite{ye2024mm} and EgoLife~\cite{yang2025egolife} collect large-scale egocentric vision data paired with text and develop foundation models for egocentric question answering and personal assistance.
Several studies~\cite{engel2023project,bao2023can,cheng2024egothink,yan2025teleego,zhou2025egotextvqa} also develop diverse datasets and benchmarks for egocentric video assistants.
Beyond foundation models and datasets, recent work has developed real-world systems that integrate diverse sensor data from mobile and wearable platforms, such as smart glasses, for everyday use.
Vinci~\cite{huang2025vinci} leverages egocentric video to provide real-time responses to user queries based on its observations and historical context.
agentAR~\cite{zhu2025agentar} develops a personal agent with an AR authoring system that integrates external tools and LLM-based reasoning to support personal question-answering tasks.
OS-1~\cite{xu2024can} is a companion on smart glasses that uses visual and audio cues from the environment to deliver personalized responses.
Several studies have explored egocentric assistants for broader applications, such as assisting individuals with visual impairments~\cite{tokmurziyev2025llm,yang2024viassist} and supporting for social interaction~\cite{wang2025egosocial,zhou2026exploring}.
While these studies primarily focus on question-answering scenarios that require explicit user instructions, \workname~aims to proactively provide assistance throughout procedural tasks based on multimodal egocentric sensor data from smart glasses.

\subsection{Continual and Procedural Personal Assistants}
Recent studies, such as Gemini~\cite{gemini}, utilize understanding capabilities of MLLMs in live visual perception to serve as personal assistants that provide real-time daily support.
However, these systems require users to explicitly specify their task progress or current step to trigger guidance for subsequent steps.
PrISM-Q\&A~\cite{arakawa2024prism} is a voice assistant that uses audio and motion sensors for human activity recognition~(HAR) and provides step-aware support for procedural tasks.
However, it remains constrained to a question-answering paradigm, requiring users to explicitly ask for guidance during procedural tasks.
\revision{Building on HAR-based step tracking, PrISM-Observer~\cite{arakawa2024prismobserver} enables proactive intervention by predicting intervention moments for a set of pre-selected steps through modeling step durations and transitions.
The PrISM framework~\cite{arakawa2025scaling} further addresses HAR's sensing imperfections through extracting step context from continuous mixed-initiative dialogue (e.g., reactive Q\&A, user self-narration, and system reminders).}
\revision{Satori~\cite{li2025satori} forecasts next-step assistance in parallel while detecting step-completion checkpoints to trigger delivery of the cached assistance through user confirmation.}
\revision{These systems trigger assistance based on discrete events (e.g., predicted time-to-step, step completion detection) and primarily deliver pre-selected or pre-forecast content for the next step.}
\revision{In contrast, \workname~exploits multimodal data from smartglasses to continuously track fine-grained task progress and reason over the user's evolving state, deciding at each moment whether and what assistance to deliver based on the user's actual state.}

\input{tables/related_work_v1}
\subsection{Proactive Assistant Systems}
Recent studies, such as VideoLLM-online~\cite{chen2024videollm}, explore enabling VLMs to support online interaction over streaming video by introducing training objectives that allow the model to determine when to respond while processing dense video sequences.
However, this line of work focuses on efficient processing of visual streams at the VLM level and does not explicitly model user intent, which is essential for providing appropriate assistance.
Beyond model-level designs, recent research has proposed proactive systems that model user intent to anticipate user needs and provide assistance without explicit user instructions~\cite{liu2024chainstream,yang2025proagent,pu2025promemassist,yang2025contextagent}.
Studies such as SocialMind~\cite{yang2025socialmind} and LLAMAPIE~\cite{chen2025llamapie} provide proactive suggestions on AR glasses or earphones during face-to-face conversations, with a primary focus on social scenarios.
ContextAgent~\cite{yang2025contextagent}, ProAgent~\cite{yang2025proagent}, and ChainStream~\cite{liu2024chainstream} can harness multimodal sensor data for reasoning and automatically decide when and what to proactively assist users.
ProMemAssist~\cite{pu2025promemassist} focuses on modeling the assistance value and interruption cost, enabling more selective proactive support during ongoing tasks.
However, these studies mainly focus on delivering one-shot proactive support for short-term events and isolated moments.
In contrast, \workname~provides continuous, step-aware assistance by incorporating multi-scale temporal procedural context and expert knowledge into reasoning, rather than relying on single-moment holistic scene understanding.

%% file: tables/related_work_v1.tex
\begin{table}[t]\footnotesize
\setlength{\tabcolsep}{3pt}

  \caption{A summary of recent egocentric assistive systems (\CIRCLE \ means included. `V' and `A' denote vision and audio).}
  \vspace{-1em}
  \label{tab:Comparison}
  \begin{tabular}{ccccccccc}
    \toprule
    \textbf{Approach} & \makecell{\textbf{Adaptive}\\ \textbf{Perception}} & \makecell{\textbf{Procedural}\\ \textbf{Tasks}} & \makecell{\textbf{Step-Aware}\\\textbf{Assistance}}  &
        \makecell{\textbf{Expert}\\ \textbf{Knowledge}} & \makecell{\textbf{Sensor}\\ \textbf{Modalities}} &
    \makecell{\textbf{Interactive}\\ \textbf{Mode}} & 
    \makecell{\textbf{Assistance}\\ \textbf{Mode}} & \makecell{\textbf{System}\\ \textbf{Settings}} \\
    \midrule
    MM-Ego~\cite{ye2024mm}& \Circle  & \Circle & \Circle &  \Circle & V& Reactive&One-shot&N.A.\\
    Vinci~\cite{huang2025vinci}& \Circle  & \CIRCLE & \CIRCLE & \Circle & V& Reactive&Continuous&Glasses\\
    agentAR~\cite{zhu2025agentar}& \Circle  & \CIRCLE & \CIRCLE & \Circle & V& Reactive&Continuous&Glasses\\
    PrISM-Q\&A~\cite{arakawa2024prism}& \Circle  & \CIRCLE & \CIRCLE & \CIRCLE & A, IMU& Reactive&Continuous&Smartwatch\\
    \revision{PrISM-Observer~\cite{arakawa2024prismobserver}}& \Circle  & \CIRCLE & \CIRCLE & \Circle & A, IMU& Proactive&Continuous&Smartwatch\\
    \revision{PrISM~\cite{arakawa2025scaling}}& \Circle  & \CIRCLE & \CIRCLE & \CIRCLE & A, IMU& Mixed&Continuous&Smartwatch\\
    VideoLLM-online \cite{chen2024videollm}& \Circle  & \Circle & \Circle & \Circle & V& Proactive&Continuous&N.A.\\
    ContextAgent \cite{yang2025contextagent}& \Circle  & \Circle & \Circle & \Circle & V& Proactive&One-shot&N.A.\\
    ProAgent \cite{yang2025proagent}& \CIRCLE & \Circle & \Circle & \Circle & V, A, IMU, GPS& Proactive&One-shot&Glasses\\    
    SocialMind \cite{yang2025socialmind}& \Circle  & \Circle & \Circle & \Circle & V, A, IMU& Proactive&Continuous&Glasses\\
    OS-1 \cite{xu2024can}& \Circle  & \Circle & \Circle & \Circle & V, A&Proactive&One-shot&Glasses\\
    \textbf{\workname~}& \CIRCLE  & \CIRCLE & \CIRCLE & \CIRCLE & V, A, IMU& Proactive&Continuous&Glasses\\
  \bottomrule
\end{tabular}
\end{table}

%% file: section/Motivation.tex
\section{BACKGROUND AND MOTIVATION}
\label{sec:motivation}

\revision{In this section, we present observations and measurements from self-collected egocentric recordings to motivate the design of \workname, using the procedural task of tea making as an example.}

\noindent\textbf{{Observation 1: \revision{Continuous, Step-Aware} Assistance in Procedural Tasks.}}
\label{sec:motivation_assist_procedural}
\revision{Unlike general scenarios where one-shot suggestions for isolated short-term events are sufficient~\cite{yang2025proagent,yang2025contextagent}, procedural tasks consist of ordered, interdependent steps whose assistance should continuously reason over the user's current step and execution status throughout the task.}
As shown in Figure~\ref{fig:motivation_proagent}, when the user is about to finish checking the water temperature during tea making, the one-shot proactive system correctly recognizes the kitchen scene but fails to deliver appropriate assistance due to inaccurate step and status recognition.
Effective proactive assistance in procedural tasks should provide step-specific instructions at the beginning of an action and suggest the next step as the current action nears completion.
\revision{Therefore, continuous, step-aware assistance is essential for procedural tasks, motivating a design that continuously reasons over the user's evolving state across interdependent steps to deliver timely guidance grounded in the user's actual state rather than one-shot suggestions.}

\input{insert_figures/motivation_headmotion}
\input{insert_figures/motivation_handmotion}

\noindent\textbf{{Observation 2: User's Intent and Attention in Procedural Task Execution.}}
\label{sec:motivation_user_intent}
Proactive assistance requires an accurate understanding of user intent. In particular, correctly identifying the user’s attentional focus is essential for providing timely and relevant assistance in procedural tasks.
\revision{
We observe that head and hand motions captured by AR glasses serve as key indicators.
On the one hand, head motion is a strong indicator of attention transitions~\cite{doshi2012head,doshi2009roles}.
As shown in Figure~\ref{fig:imu_design}, when the user transitions to the step of measuring cold water and searches for the water bottle while making tea, head motion exhibits significant movement patterns.
In contrast, when users focus on a specific manipulation, like pouring water into the measuring cup, head motion remains stable.
On the other hand, hand manipulation in egocentric vision is critical for inferring user intent and execution status~\cite{bandini2020analysis,EgoProceLECCV2022,lee2024error}.}
As shown in Figure~\ref{fig:hand_motion_cues}, hand manipulation directly reflects the user’s focus and step execution status, providing rich temporal information about short-term procedural dynamics.
\revision{Therefore, head and hand motions serve as cues of user intent and attention in procedural tasks, motivating our use of egocentric multimodal data from AR glasses to capture these signals.}

\noindent\textbf{{Observation 3: Action Understanding for Procedural Tasks.} }
\label{sec:mitivation_action_understanding}
As shown in Figure~\ref{fig:motivation_au_examples}, action understanding in procedural tasks exhibits two characteristics.
\revision{First, visually similar but functionally opposite manipulations within the same step, such as ``tilting a cup to pour water’’ versus ``lifting the cup after pouring'' in Figure~\ref{fig:motivation_au_examples} (a), should be distinguished to identify execution status.
Distinguishing them benefits from short-term hand manipulation cues that capture temporal dynamics within a step.
Second, similar actions belonging to different steps, such as ``placing a tea bag into a mug'' versus ``steeping the tea bag''  in Figure~\ref{fig:motivation_au_examples} (b), share similar visual appearance but correspond to different procedural stages.
Distinguishing them benefits from both task-specific expert knowledge, which provides step ordering, dependencies, and step-level execution details, and long-term task progress, which tracks completed steps throughout task execution. Together, these two sources offer complementary information that helps better understand the user's ongoing action.
For example, knowing that ``steeping'' must follow ``placing the tea bag'' and that ``placing'' has already been completed allows the system to correctly identify the current action as ``steeping’’.}
We adapt existing VLMs of varying scales to procedural action understanding using in-context learning~\cite{dong2024survey} with five examples.
As shown in Figure~\ref{fig:motivation_vlms}, existing VLMs struggle to identify both the step and its execution status when relying solely on intrinsic knowledge and single-moment reasoning, as procedural tasks are inherently sequential and interdependent.
Meanwhile, as shown in Figure~\ref{fig:motivation_overhead}, incorporating temporal context directly through dense frame sequences incurs rapidly growing computational overhead, posing challenges for real-time assistance.
\revision{Therefore, both short-term and long-term temporal context and task-specific expert knowledge are essential for effective procedural action understanding, motivating a design that integrates procedural knowledge with efficient temporal modeling for real-time assistance.}
\input{insert_figures/motivation_handmotion_v2}

\input{insert_figures/motivation_action_understanding}

%% file: insert_figures/motivation_headmotion.tex
\begin{figure*}[t]
\small
\centering
\begin{minipage}[t]{0.4\textwidth} %
\centering
\includegraphics[width=\textwidth]{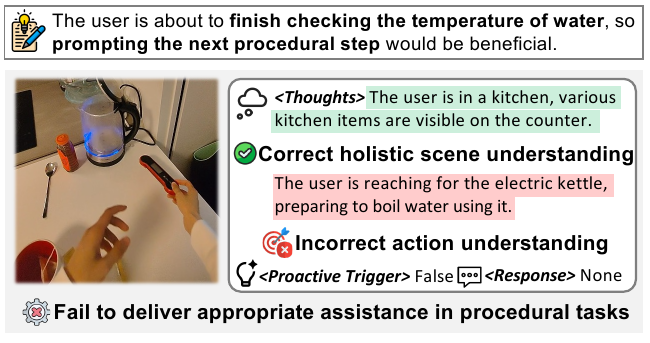}
\captionsetup{skip=3pt}
\caption{An example of an existing proactive system in procedural tasks~\cite{yang2025proagent}.}
    \label{fig:motivation_proagent}
\end{minipage}
\hfill
\begin{minipage}[t]{0.59\textwidth} %
\centering
\includegraphics[width=\textwidth]{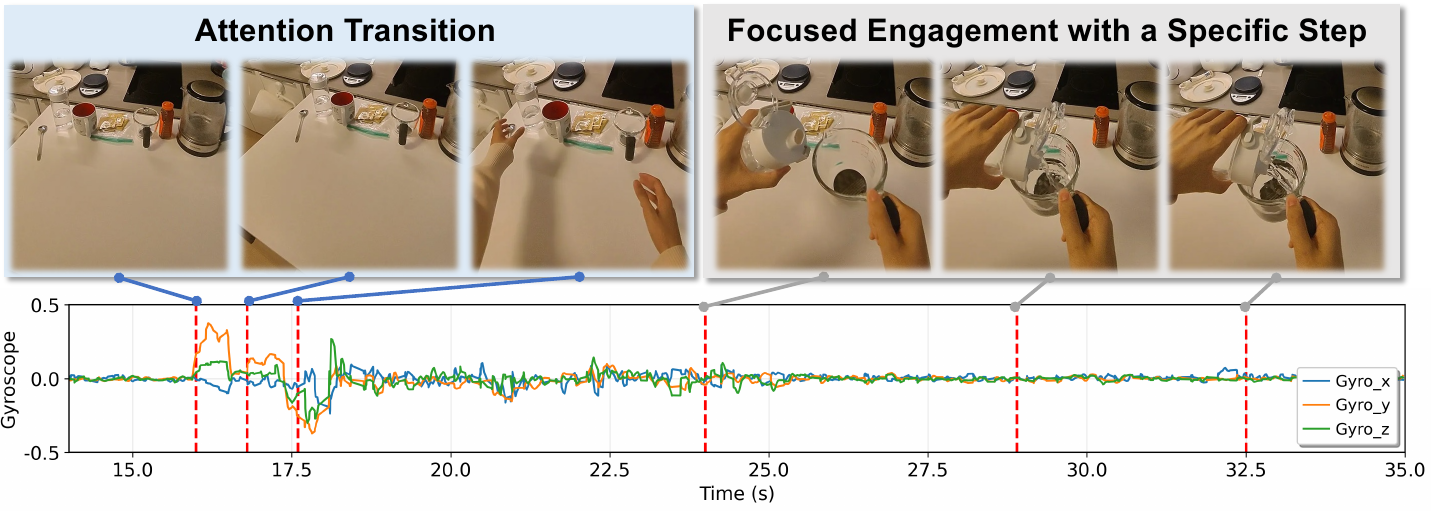}
\captionsetup{skip=3pt}
    \caption{
    A preliminary example that head motion can effectively indicate attention transitions during procedural task execution.
    }
    \label{fig:imu_design}
\end{minipage}
\end{figure*}

%% file: insert_figures/motivation_handmotion.tex
\begin{figure}[t]
    \centering
    \captionsetup{skip=3pt}
    \includegraphics[width=\textwidth]{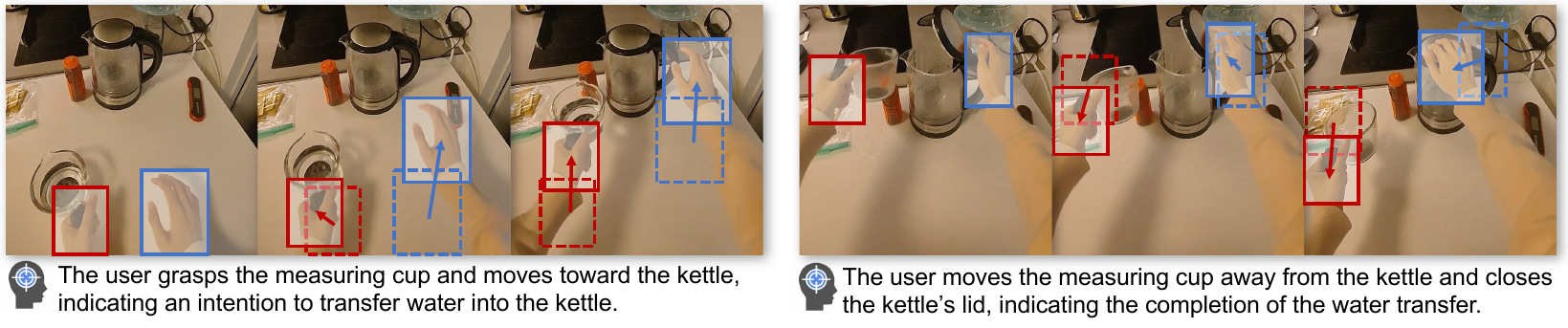}
    \caption{Egocentric vision and hand motion provide important cues for inferring user intent during procedural tasks.}
    \label{fig:hand_motion_cues}
\end{figure}

%% file: insert_figures/motivation_handmotion_v2.tex
\begin{figure}[t]
    \centering
    \captionsetup{skip=3pt}
    \includegraphics[width=\textwidth]{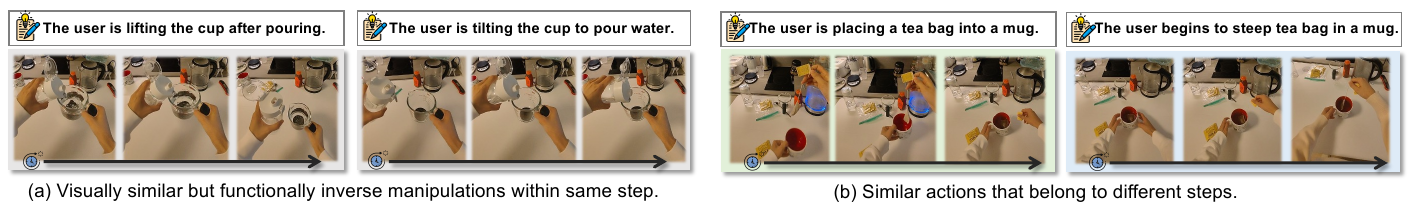}
    \caption{Examples illustrating two characteristics in action understanding for procedural tasks: visually similar but functionally inverse manipulations within the same step (left), and similar actions that correspond to different steps (right). }
    \label{fig:motivation_au_examples}
\end{figure}

%% file: insert_figures/motivation_action_understanding.tex
\begin{figure*}[t]
\small
\centering
\begin{minipage}[t]{0.49\textwidth} %
\centering
 \includegraphics[width=\textwidth]{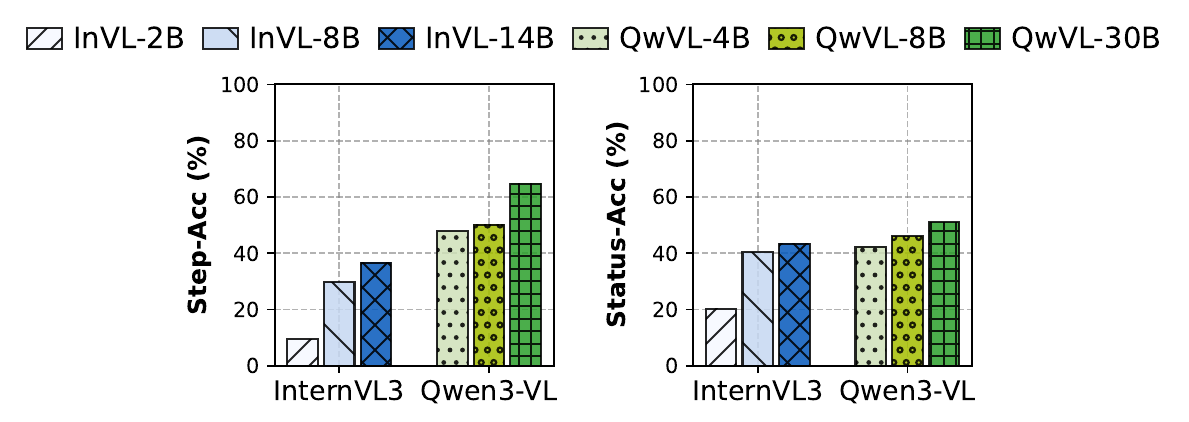}
\captionsetup{skip=1pt}
\caption{Performance of existing VLMs adapted for procedural action understanding. ``Step-Acc'' and ``Status-Acc'' denote step and execution status identification accuracy, respectively.}
    \label{fig:motivation_vlms}
\end{minipage}
\hfill
\begin{minipage}[t]{0.49\textwidth} %
\centering
\includegraphics[width=\textwidth]{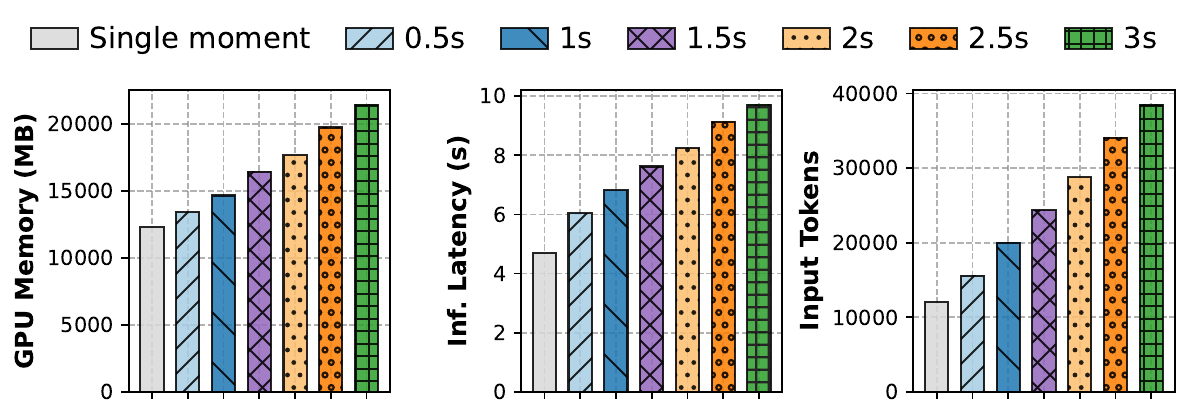}
\captionsetup{skip=1pt}
\caption{System overhead increases significantly as more input frames are incorporated as temporal context. ``Inf.'' denotes inference.}
    \label{fig:motivation_overhead}
\end{minipage}
\end{figure*}

%% file: section/System_design.tex
\section{SYSTEM DESIGN}

\subsection{System Overview}
\workname~is an end-to-end assistive system that exploits multimodal egocentric sensor data from AR glasses to provide continuous, step-aware assistance during procedural tasks.
Figure \ref{fig:system_overview} demonstrates the overview of \workname, which consists of three major modules.
First, \workname~performs \textit{\revision{motion-based} perception}~(\S~\ref{sec:attention-based perception}) by adaptively sampling visual data using a head-motion-aware strategy and extracting key moments via motion-based selection.
This design utilizes attention cues from multimodal egocentric data to reduce redundant visual processing while preserving moments with high potential for proactive needs.
Second, \workname~achieves \textit{step-oriented procedural context extraction}~(\S~\ref{sec:context-construction}) by effectively incorporating task-specific expert knowledge about the particular procedural
task and multi-scale temporal context, including short-term hand motion cues and long-term task progress. 
The extracted context captures both procedural knowledge and temporal dynamics, which are important for procedural action understanding.
Finally, \textit{step-aware proactive reasoner}~(\S~\ref{sec:proactive-reasoner}) in \workname~integrates sensory context with the extracted procedural context to perform motion-aware action understanding and step-aware proactive reasoning.
Proactive responses are generated only when necessary, and a step-aware consistency checking mechanism is introduced to suppress redundant feedback and mitigate single-moment mispredictions by leveraging temporal consistency across historical predictions.
These responses are then displayed on the AR glasses, allowing users to keep the task scene in view while receiving timely assistance.

\input{insert_figures/system_overview}

\subsection{\revision{Motion-Based} Perception}
\label{sec:attention-based perception}

Unlike reactive assistants~\cite{huang2025vinci,xu2024can} triggered by explicit queries, a proactive assistant must continuously observe user state to identify moments requiring assistance.
However, such moments are typically sparse in long egocentric streams, making uniform processing with intent inference inefficient and unnecessary. 
\revision{As discussed in \S\ref{sec:motivation_user_intent} (Observation 2), motion cues from multimodal egocentric sensing can indicate user intent and attention.
Thus, \workname~introduces a motion-based perception mechanism for efficient intent inference, leveraging \final{always-on, low-cost} head motion signals to adaptively guide visual sampling and \final{fine-grained visual} motion cues to \final{further} identify moments with high potential for proactive needs from continuous observations.}

\subsubsection{Head-Motion-Aware Adaptive Sampling}
\revision{To efficiently decide \final{how densely} to capture frames, \workname~exploits head motion from IMU as an always-on sampling signal to drive adaptive vision sampling, as it is far cheaper than continuous vision processing, while effectively reflecting potential attention transitions during procedural tasks.}
Specifically, \workname~measures head motion \final{as the magnitude of the} angular velocity measured by the gyroscope, \revision{which captures rotational head movements with low cost and latency}.
\workname~continuously monitors head motion and applies a motion-aware sampling threshold $\tau_{\text{s}}$ to determine the visual data sampling rate.
When head motion remains below $\tau_{\text{s}}$, indicating stable head orientation, the system operates in a normal sampling mode,
\final{which still periodically captures visual observations of ongoing task execution, including moments when users focus on specific manipulations within a step}.
Once head motion exceeds $\tau_{\text{s}}$, suggesting a potential attention shift (e.g., navigating between steps or searching for objects), the system temporarily switches to a burst sampling mode to capture visual information more densely around these moments.
In practice, the normal and burst sampling modes trigger a frame pair every $1\,\text{s}$ and $0.5\,\text{s}$, respectively, with each pair consisting of two consecutive frames with a $0.1\,\text{s}$ interval for optical flow computation (\S\ref{sec:moment_selection}).
\revision{This design converts always-on low-cost sensing into a motion-aware vision capture strategy} that densely samples around critical moments that may require proactive assistance while reducing sampling overhead during stable periods.

\subsubsection{Motion-Based Key Moment Selection}
\label{sec:moment_selection}
\revision{Unlike anomaly detection on sensor signals that filters idle moments for improving activity recognition~\cite{arakawa2025scaling}, \workname~uses \final{visual} motion cues to identify frames that need VLM reasoning for efficient intent inference.}
Even with adaptive sampling, not all captured frames are equally informative for proactive reasoning.
\revision{We therefore leverage optical flow, which provides dense motion cues, to further identify frames with meaningful contextual changes.
A direct approach would compute optical flow uniformly over the full frame.
However, our observations in \S\ref{sec:motivation_user_intent} show that hand manipulations are particularly informative of user attention and execution status, which can further provide essential hand motion cues for downstream VLM reasoning (see \S~\ref{sec:temporal-context} for details).}
Meanwhile, as shown in Figure~\ref{fig:HRA_design}, optical flow cost scales with region size, and hand regions typically occupy less than 30\% of the frame, indicating that restricting optical flow computation to hand-related regions can effectively reduce processing overhead.
\revision{Thus, as shown in Figure~\ref{fig:opticalflow_design}, \workname~first detects hands in the field of view, then computes optical flow over hand regions if hands are detected, or over the full frame otherwise.}
Frames with optical flow magnitude below a motion-based filtering threshold $\tau_{\text{f}}$ are filtered out to avoid redundant intent inference, while the remaining frames serve as key moments suggesting a higher likelihood of proactive assistance needs.
By focusing on these motion-rich frames, \workname~reduces the processing burden on subsequent VLM reasoning while retaining moments where proactive assistance is needed.

\input{insert_figures/deign_optical}

\subsection{Step-Oriented Procedural Context Extraction}
\label{sec:context-construction}

\revision{As discussed in §\ref{sec:mitivation_action_understanding} (Observation 3), effective procedural action understanding relies on both temporal context and task-specific expert knowledge.
Thus, \workname~proposes a step-oriented procedural context extraction approach for effective modeling of procedural knowledge and temporal dynamics, enabling continuous tracking of fine-grained task progress.}

\subsubsection{Expert Knowledge Retrieval}
\label{sec:guideline}
Many multi-step physical tasks in daily life, from cooking to assembly, rely on task-specific expert knowledge that is typically provided through instruction guidelines~\cite{aggarwal2025generating}.
The right side of Figure~\ref{fig:motivation_stepid} illustrates a step graph derived from a procedural guideline that explicitly specifies both sequential and parallel dependencies between procedural steps and step-level execution details.
\revision{These dependencies and execution details together form the expert knowledge that \workname~integrates to better infer the user's current task state and provide informative, contextually appropriate guidance.}
Specifically, \workname~takes an initial task instruction $\mathcal{I}$ as input to initiate the service.
This instruction could be explicitly provided by the user through speech (e.g., ``I'm going to brew a cup of tea. Please provide step-by-step guidance.''), as in our work, or implicitly inferred by a one-shot proactive system~\cite{yang2025contextagent} from environmental context.
Next, \workname~retrieves a guideline $g_{task}$ from WikiHow~\cite{koupaee2018wikihow}, a large-scale public repository of procedural instruction articles in free-form natural language, i.e., $g_{task} = \texttt{Retriever}(\mathcal{I}, \mathcal{D}_{task})$, where \texttt{Retriever} is the retrieval model.
\revision{However, directly feeding $g_{task}$ into the system introduces an additional burden during real-time inference, as free-form guidelines leave inter-step dependencies implicit and would require the system to extract structured procedural knowledge from free-form text at every reasoning turn.
\workname~therefore offloads this extraction to an advanced LLM (e.g., Claude or GPT series~\cite{singh2025openai,Claude-Sonnet-4.5}), whose strong language capabilities have been widely validated, as a one-time step at task initiation.} 
The LLM extracts task-specific expert knowledge from the retrieved guideline and transforms it into a structured representation $\mathcal{G}=\text{LLM}(p, \textit{examples}, g_{task})$, where $p$ denotes the prompt (details in Appendix~\ref{appendix:prompts}) and \textit{examples} are few-shot guideline exemplars demonstrating the desired structured representation, constructed based on the EgoPER dataset~\cite{lee2024error}.
\workname~then incorporates the structured guideline $\mathcal{G}$ into the VLM prompt to support step-aware reasoning (details are provided in \S\ref{sec:proactive-reasoner}).

\subsubsection{Multi-Scale Temporal Context Extraction}
\label{sec:temporal-context}
As shown in Figures~\ref{fig:motivation_status} and~\ref{fig:motivation_stepid}, relying on single-moment observations makes it difficult to distinguish between different execution statuses within the same step and between similar actions of different steps, indicating that effective procedural action understanding and continuous tracking require temporal context.
\revision{Such temporal context includes short-term hand manipulations that capture dynamics within a step, and long-term task progress that tracks dynamics across the task workflow.
Rather than directly passing dense visual streams to the VLM, which introduces rapidly growing computational overhead~(Figure~\ref{fig:motivation_overhead}), \workname~extracts each scale through compact textual cues for efficient temporal modeling.}

\input{insert_figures/motivation_stepid}
\noindent\textbf{{Short-term Hand Motion Cues.}}
\revision{The optical flow estimated during motion-aware perception (\S\ref{sec:attention-based perception}) provides a natural source of short-term motion cues. A straightforward approach would be to feed optical flow visualizations directly into VLMs.}
However, existing VLMs typically struggle to interpret optical flow representations directly, as they are primarily trained on RGB data~\cite{schuhmann2022laion5b, lin2015coco}.
Moreover, dense per-pixel motion information are redundant for procedural action understanding, which is mainly driven by hand manipulations.
Therefore, \workname~extracts motion cues from hand-related regions and represents them in a compact textual form.
As shown on the left side of Figure~\ref{fig:opticalflow_design}, optical flow is estimated within hand-related regions between consecutive frames.
Per-pixel flow vectors are then aggregated and decomposed into motion magnitude and direction angle, with the direction angle mapped into one of eight cardinal directions defined on the 2D image plane~(up, down, left, right, and four diagonals).
The result is converted into natural language descriptions, such as \textit{``The left hand is moving down-right, the right hand remains almost stationary.''}
\workname~incorporates these hand motion cues into the VLM prompt as descriptions of image-plane hand movements (details are provided in \S~\ref{sec:proactive-reasoner}), enabling the VLM to better interpret hand manipulations and improve procedural action understanding, as illustrated at the bottom of Figure~\ref{fig:motivation_status}.

\noindent\textbf{Long-term Historical Task Progress.}
\revision{Incorporating past reasoning traces into the VLM prompt would provide long-term temporal context, but these traces accumulate as the task proceeds, leading to overly long prompts and degraded reasoning performance.}
\revision{\workname~therefore maintains a compact structured text record of completed steps from past verified predictions as the user performs the task, e.g., \textit{``[Measure 12 ounces of water, Transfer water to kettle]''}.
This sequence of completed steps, combined with the retrieved guideline $\mathcal{G}$, effectively indicates task progress and localize the current step. The record is initialized as empty at the start of each task session and progressively updated as verified step transitions occur throughout the task (see \S\ref{sec:consistency_checking} for the historical context update mechanism).}
As shown on the left side of Figure~\ref{fig:motivation_stepid}, combining this progress record with the retrieved guideline enables the VLM to better identify the user’s current step within the task workflow, thereby distinguishing visually similar actions across different steps and anticipating possible next steps.

\subsection{Step-Aware Proactive Reasoner}
\label{sec:proactive-reasoner}

\revision{As discussed in §\ref{sec:motivation_assist_procedural} (Observation 1), assistance for procedural tasks requires the system to reason over the user's evolving state in the task rather than perform single-moment holistic scene understanding.
\workname~therefore introduces a step-aware proactive reasoner that jointly reasons over visual input with the extracted procedural context, enabling the system to capture evolving user needs and deliver continuous, step-aware assistance throughout tasks.
A consistency checking mechanism further mitigates the impact of single-moment mispredictions and suppresses redundant responses.}

\input{insert_figures/vlm_reasoning}

\subsubsection{VLM Reasoner Training and Inference}
\revision{Unlike prior work that forecasts next-step assistance and waits for step completion triggers through two parallel processes~\cite{li2025satori}, \workname~uses a single VLM reasoner to continuously reason over the user's evolving state, deciding whether and what assistance to provide, grounded in the user's actual state at each inference moment.}
\revision{Achieving this requires the reasoner to learn not only what the user's state is, but also whether assistance is needed and how to generate responses aligned with that state.}
\revision{\workname~therefore trains the reasoner using a combination of ground-truth supervision with offline distillation from an advanced LLM that provides explicit reasoning chains, equipping it to jointly perform procedural action understanding and step-aware proactive reasoning.}

Figure~\ref{fig:vlm_reasoning} illustrates the overall training pipeline.
During training, each instance consists of multimodal inputs and hierarchical supervision targets.
The inputs consist of the egocentric image, structured guideline, and multi-scale temporal context.
The supervision targets include ground-truth step, execution status, and proactive trigger labels, along with LLM-generated motion-aware action understanding and step-aware proactive responses as distillation targets.
All LLM-generated annotations are validated by human annotators for quality and consistency (details in Appendix~\ref{appendix:annotation}).
\revision{The distillation targets equip the reasoner with two key capabilities.
For motion-aware action understanding, the advanced LLM generates explicit reasoning chains that map visual and motion cues to the corresponding step and execution status, teaching the reasoner to interpret short-term hand motion cues. 
For step-aware proactive response generation, the LLM generates high-quality step-aware responses, demonstrating how to align guidance with the user's current state and procedural knowledge.} 
Through supervised training with distillation, the VLM reasoner learns to reproduce both behaviors while predicting ground-truth labels.

During inference, guided by the system prompt shown in Figure~\ref{fig:system_prompt}, the reasoner identifies the step with execution status and determines whether assistance is needed.
If no assistance is required, it remains silent to avoid unnecessary interruption. Otherwise, it generates a step-aware proactive response aligned with the user’s state to guide them in completing the current step or proceeding to the next.

\input{insert_figures/system_prompt}

\subsubsection{Step-Aware Consistency Checking}
\label{sec:consistency_checking}

While \workname~leverages temporal context for \revision{continuous} reasoning rather than treating each prediction independently, single-moment mispredictions can occur and accumulate over time.
Moreover, repeatedly delivering similar responses for the same user state can be disruptive and distract the user from the task at hand.
\revision{To ensure reliable and non-intrusive assistance, as shown in Algorithm~\ref{alg:consistency_checking}, \workname~proposes a step-aware consistency checking mechanism that handles the reasoner's predictions, benefiting both long-term temporal context update and response delivery.}

\noindent\textbf{{Historical Context Update Mechanism.}}
Maintaining long-term historical task progress based on past observations and predictions is critical for continuous reasoning.
However, directly incorporating every single-moment prediction into the historical record would propagate mispredictions and degrade subsequent reasoning.
\revision{Unlike passively decaying memory items based on temporal scores~\cite{pu2025promemassist}, \workname~therefore uses a sliding window as a temporal consistency check over historical step predictions, updating the record only when step transitions are consistently observed across multiple moments.}
\revision{Specifically, when the predicted step $\hat{D}_s$ differs from the active step $D_{s}^{\mathrm{act}}$, \workname~treats this as a candidate transition rather than immediately updating the historical record~$\hat{H}_p$.
It then evaluates predictions within the sliding window $\mathcal{B}$, adding $D_{s}^{\mathrm{act}}$ to the record as completed only when the new step appears in the majority of $\mathcal{B}$.}
This update mechanism mitigates the impact of single-moment mispredictions and ensures that only temporally consistent step transitions are reflected in the historical task progress, providing more reliable long-term context for subsequent reasoning.

\noindent\textbf{Response Delivery Control.}
Users often remain in the same step and execution status for some time.
Repeating similar proactive assistance under such stable states provides little additional benefit and could distract the user from the ongoing task.
To avoid repetitive interruptions, \revision{\workname~suppresses response delivery when the predicted step and execution status ($\hat{D}_s$, $\hat{S}_s$) match those from the previous moment ($D_{s}^{\mathrm{prev}}$, $S_{s}^{\mathrm{prev}}$), indicating that the user is during steady execution and does not need further intervention.}
This mechanism mitigates redundant responses and ensures that proactive assistance is delivered when meaningful state changes occur.

%% file: insert_figures/system_overview.tex
\begin{figure}[t]
    \centering
    \captionsetup{skip=5pt}
    \includegraphics[width=\textwidth]{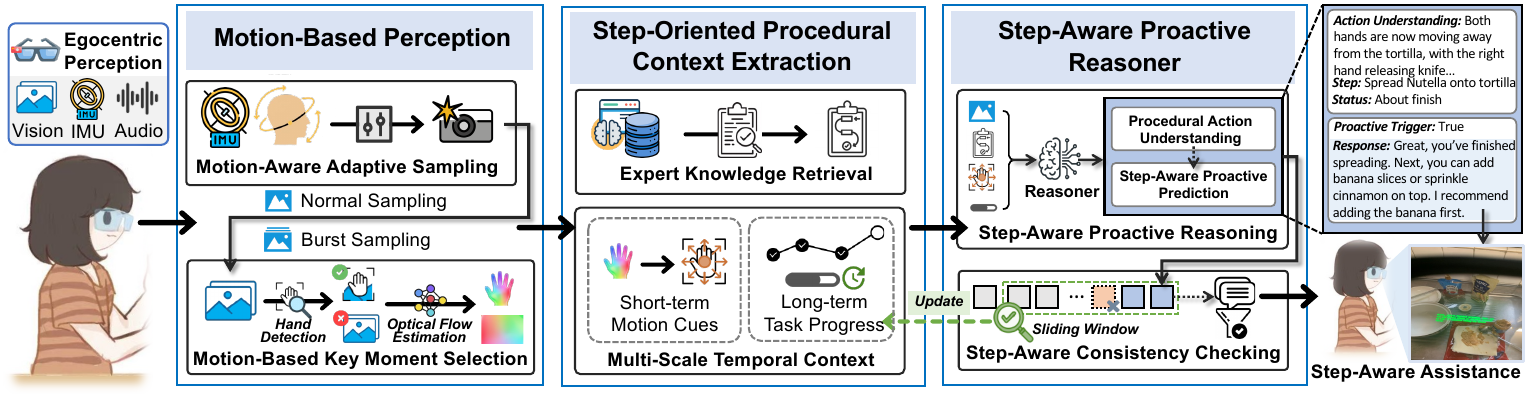}
    \caption{System overview of \workname.
    \workname~utilizes multimodal egocentric data from AR glasses to achieve \revision{motion-based} perception. 
    By integrating visual inputs with extracted task-specific expert knowledge and multi-scale temporal context, the reasoner performs step-aware proactive reasoning with consistency checking. The resulting proactive assistance is then delivered to the user via on-screen displays on the AR glasses. %
    }
    \label{fig:system_overview}
\end{figure}

%% file: insert_figures/deign_optical.tex
\begin{figure*}[t]
\small
\centering
\begin{minipage}[t]{0.49\textwidth} %
 \includegraphics[width=\textwidth]{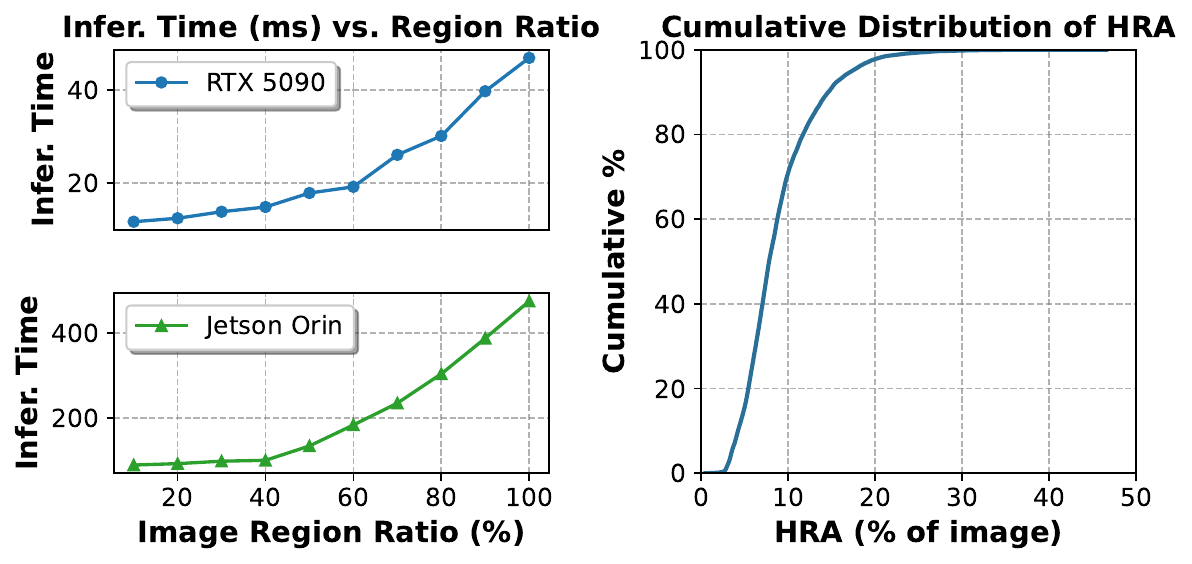}
 \captionsetup{skip=5pt}
\caption{Optical flow computation time and area ratio distribution of hand-related regions~(HRA).}
\label{fig:HRA_design}
\end{minipage}%
\hfill
\begin{minipage}[t]{0.49\textwidth} %
\centering
 \includegraphics[width=\textwidth]{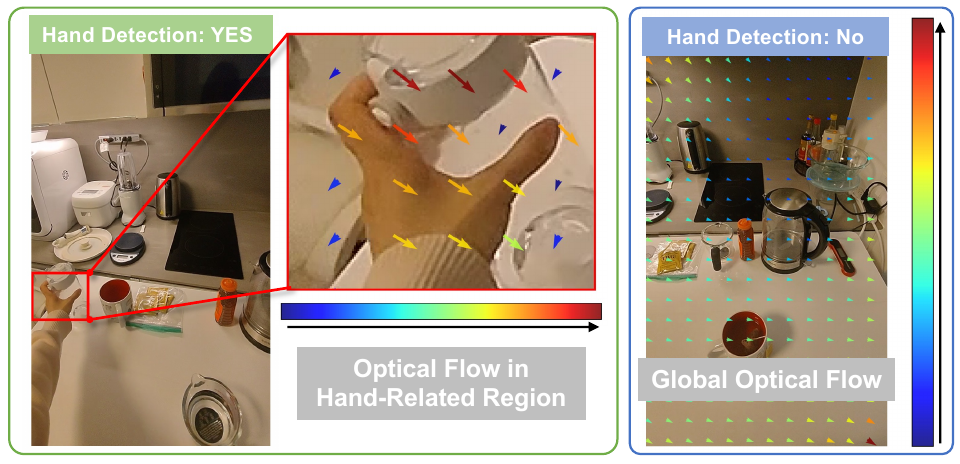}
 \captionsetup{skip=5pt}
    \caption{Optical flow estimation adapts to hand presence. Arrows and colors represent motion direction and magnitude.}
    \label{fig:opticalflow_design}
\end{minipage}

\end{figure*}

%% file: insert_figures/motivation_stepid.tex
\begin{figure*}[t]
\small
\centering
\begin{minipage}[t]{0.345\textwidth} %
\centering
 \includegraphics[width=\textwidth]{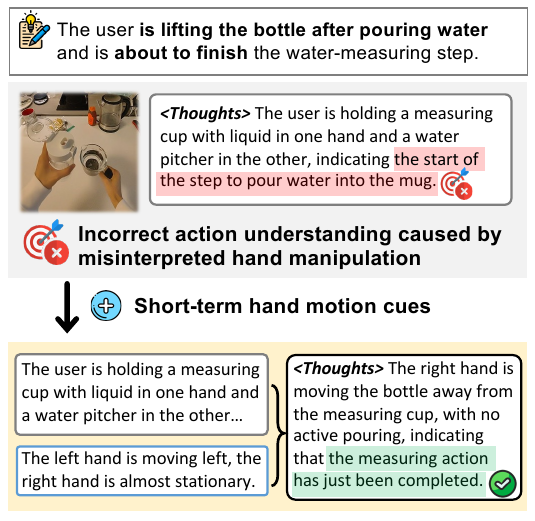}
 \captionsetup{skip=5pt}
    \caption{Impact of incorporating short-term hand motion cues and their impact.}
    \label{fig:motivation_status}
\end{minipage}%
\hfill
\begin{minipage}[t]{0.645\textwidth} %
\centering
  \includegraphics[width=\textwidth]{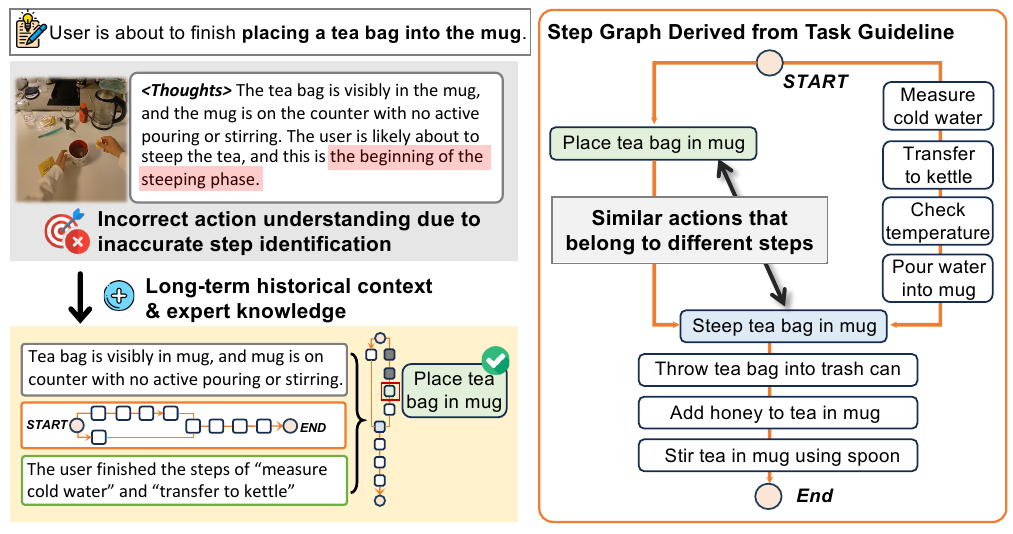}
  \captionsetup{skip=5pt}
    \caption{An Illustration of incorporating long-term historical task progress and task-specific expert knowledge.}

    \label{fig:motivation_stepid}
\end{minipage}

\end{figure*}

%% file: insert_figures/vlm_reasoning.tex
\begin{figure}[t]
    \centering
    \captionsetup{skip=5pt}
    \includegraphics[width=0.59\textwidth]{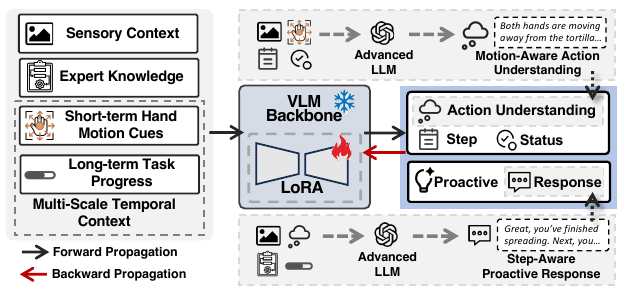}
    \caption{Training pipeline of the step-aware proactive reasoner in \workname.}
    \label{fig:vlm_reasoning}
\end{figure}

%% file: insert_figures/system_prompt.tex
\begin{figure}[t]
\begin{minipage}[t]{0.48\textwidth}
\centering
\begin{tcolorbox}[
    enhanced,
    colback=white,
    colframe=black!60,
    boxrule=0.6pt,
    arc=0mm,
    left=1.2mm,right=1.2mm,top=1.0mm,bottom=1.0mm,
    title=\textbf{Template of System Prompt},
    coltitle=black,
    colbacktitle=gray!20,
    fonttitle=\footnotesize \bfseries,
    halign title=center
]
\footnotesize 
\textbf{Task Instructions:} You are a proactive assistant for procedural tasks, tracking user progress through real-time sensory data from wearable devices and procedural context. Based on these contexts, you need to understand user's actions and identify the current procedural step and status, then determine whether to initiate a proactive service for the user or not. Please provide timely, relevant guidance when appropriate.\\[5pt]
Now you will receive the sensory and procedural contexts for the target moment in the procedural task.\\
\textbf{Guideline:} Structured guideline with task-specific expert knowledge: \textit{$\langle$GUIDELINE$\rangle$}\\
\textbf{Historical Context:} Previously completed steps in order: \textit{$\langle$STEP-1$\rangle$}, \textit{$\langle$STEP-2$\rangle$},\ldots\\
\textbf{Sensory Context:} Vision information shows as the image. \textit{$\langle$IMAGE$\rangle$}\\
\textbf{Hand Motion Cues:} The optical flow between the current frame and the previous frame indicates that in the image plane, \textit{$\langle$HAND\_MOTION\_CUES$\rangle$}. You may combine the image-plane hand motion cue with visual context, object relationships, and task semantics to infer how the hand is moving in the actual scene when useful.
\end{tcolorbox}
\captionsetup{skip=2pt}
\caption{System prompt for the step-aware proactive reasoner in \workname.}
\label{fig:system_prompt}
\end{minipage}%
\hfill
\begin{minipage}[t]{0.48\textwidth}
\centering
\begin{tcolorbox}
[
    enhanced,
    colback=white,
    colframe=black!60,
    boxrule=0.6pt,
    arc=0mm,
    left=1.2mm,right=1.2mm,top=1.0mm,bottom=1.0mm,
    coltitle=black,
    colbacktitle=gray!20,
    fonttitle=\footnotesize\bfseries,
    halign title=center
]
\footnotesize 
\revision{\textbf{Input:} Predicted step $\hat{D}_s$, execution status $\hat{S}_s$, proactive trigger $\hat{P}_l$, and proactive response $\hat{P}_r$ at moment $t$\\
\textbf{State:} Historical task progress $\hat{H}_p$ (an ordered list, initially empty); active step $D_{s}^{\mathrm{act}}$ (initially the first predicted step); sliding window $\mathcal{B}$ (max size $W$); previously predicted step $D_{s}^{\mathrm{prev}}$ and status $S_{s}^{\mathrm{prev}}$ (both initially empty)\\[2pt]
\textit{// Mechanism 1: Historical Context Update}\\
1:\; Append $\hat{D}_s$ to $\mathcal{B}$; remove oldest if $|\mathcal{B}| > W$\\
2:\; $D_{s}^{\mathrm{dom}} \gets$ most frequent step in $\mathcal{B}$\\
3:\; \textbf{if} $D_{s}^{\mathrm{dom}} \neq D_{s}^{\mathrm{act}}$ \textbf{and} $\mathrm{count}(D_{s}^{\mathrm{dom}}, \mathcal{B}) \geq \tau_c \cdot |\mathcal{B}|$ \textbf{then}\\
4:\;\quad Append $D_{s}^{\mathrm{act}}$ to $\hat{H}_p$ \hfill $\triangleright$ mark completed\\
5:\;\quad $D_{s}^{\mathrm{act}} \gets D_{s}^{\mathrm{dom}}$\\[2pt]
\textit{// Mechanism 2: Response Delivery Control}\\
6:\; \textbf{if} $\hat{P}_l = \mathrm{true}$ \textbf{then}\\
7:\;\quad \textbf{if} $\hat{D}_s = D_{s}^{\mathrm{prev}}$ \textbf{and} $\hat{S}_s = S_{s}^{\mathrm{prev}}$ \textbf{then}\\
8:\;\quad\quad $\hat{P}_r \gets \mathrm{None}$ \hfill $\triangleright$ suppress redundant\\
9:\;\quad \textbf{else}\\
10:\;\quad\quad Deliver $\hat{P}_r$\hfill $\triangleright$ display response\\
11:\; \textbf{else}\\
12:\;\quad $\hat{P}_r \gets \mathrm{None}$ \hfill $\triangleright$ no proactive need\\
13:\; $D_{s}^{\mathrm{prev}} \gets \hat{D}_s$;\; $S_{s}^{\mathrm{prev}} \gets \hat{S}_s$\\[2pt]
\textbf{Output:} Updated $\hat{H}_p$; $\hat{P}_r$ (or None if suppressed)}
\end{tcolorbox}
\captionsetup{skip=2pt}
\captionof{algorithm}{\revision{Step-aware consistency checking, including historical context update and response delivery control.}}
\label{alg:consistency_checking}
\end{minipage}
\end{figure}

%% file: section/Evaluation.tex
\section{EVALUATION}
In this section, we first introduce the system implementation and the datasets used for evaluation, including a dataset curated from public sources and a real-world dataset collected using our testbed.
We then introduce the evaluation metrics and baseline methods.
Finally, we present the evaluation results of \workname~together with findings from a user study.

\subsection{System Implementation}
\subsubsection{Testbed Setup.}

\revision{
We implement \workname~on a real-world hardware testbed consisting of RayNeo X3 Pro~\cite{rayneo} smart glasses and a back-end server.
The RayNeo X3 Pro integrates a Sony IMX681 camera, a built-in IMU, a four-speaker audio system, and a dual micro-LED projector system.
Its battery supports up to 5 hours of continuous operation, which is sufficient to run \workname~throughout typical procedural tasks.
A Kotlin-based Android client runs on the glasses to collect multimodal sensor data (vision, IMU, and audio), transmit it to the server over WiFi~6, and display assistance on the glasses.
\workname~is evaluated on three back-end platforms, including NVIDIA Jetson Orin and two servers with NVIDIA RTX 5090 and RTX PRO 6000 GPUs, respectively.
Lightweight models are implemented using PyTorch~\cite{paszke2019pytorch}, and we use Ollama~\cite{ollama} for VLM inference.
}

\subsubsection{Configuration.}

In the experiments, we fine-tune VLMs using low rank adaptation (LoRA)~\cite{hu2022lora} with a rank of 8, training for 10 epochs at a learning rate of $5 \times 10^{-4}$ on an NVIDIA RTX PRO 6000 GPU. We fine-tune YOLO11n~\cite{yolo11_ultralytics} on the EgoPER dataset~\cite{lee2024error} for hand detection, and employ the RAFT model~\cite{teed2020raft} for optical flow estimation.
The sampling threshold~$\tau_{\text{s}}$ and the filtering threshold $\tau_{\text{f}}$ are set to 0.3 and 10, respectively.
For expert knowledge retrieval, we use Azure Speech Recognition~\cite{microsoft_speaker_recognition} for speech-to-text conversion and \texttt{all-mpnet-base-v2}~\cite{all-mpnet-base-v2} for guideline retrieval, with embeddings of WikiHow articles precomputed and stored locally.
The advanced LLMs used in \workname~are GPT-5~\cite{singh2025openai} and Claude-Sonnet-4.5~\cite{Claude-Sonnet-4.5}.
\revision{For evaluation, we deploy \workname~with seven VLMs of varying scales, including Qwen3-VL series (2B/4B/8B/30B)~\cite{Qwen3-VL} and InternVL3 series (2B/8B/14B)~\cite{chen2024internvl}.
By default, we use Qwen3-VL-4B, and evaluate all baselines at the same scale unless otherwise specified.}

\subsection{Experimental Setup}
\subsubsection{Dataset}
\label{sec:dataset}
Although existing procedural datasets contain egocentric videos~\cite{lee2024error,li2015delving,EgoProceLECCV2022,tent}, there is no dataset containing proactive assistance annotations that can be directly used for evaluation.
Therefore, we first curate a dataset by augmenting public procedural datasets with annotations for proactive procedural assistance.
In addition, we collect a real-world dataset of multimodal data (egocentric video, IMU, and audio) using our testbed.

\noindent\textbf{Proactive Procedural Dataset.} 
We construct a dataset by augmenting public egocentric procedural data with fine-grained annotations of procedural action understanding and step-aware proactive assistance.
These annotations are generated jointly by human annotators and advanced LLMs.

\textbf{\textit{Egocentric Visual Data Source.}}
The public sources used to curate our dataset are as follows.
\begin{itemize}[leftmargin=*]
\item GTEA dataset~\cite{fathi2011learning} contains videos of seven daily activities performed by four participants, captured using a GoPro camera mounted on a cap at 30 FPS.
In experiments, we use the released rectified data provided at 15 FPS.
We utilize three relatively complex tasks with five procedural steps each (making hotdog, cheese sandwich and peanut-butter sandwich) from it, and randomly sample videos from three participants for each task.

\item EgoPER dataset~\cite{lee2024error} consists of egocentric procedural task videos with step annotations collected from 11 participants wearing Microsoft HoloLens2 at 15 FPS.
It includes five tasks (making pinwheels, quesadilla, oatmeal, coffee, and tea), with more than ten steps on average per task.
For each task, we randomly select videos from three to four participants with different execution orders.

\item EgoProceL dataset~\cite{EgoProceLECCV2022} contains egocentric videos with key-step annotations for procedural learning, constructed from both public datasets and self-collected videos, with sampling rates between 12 FPS and 60 FPS.
We utilize two daily procedural tasks (tent assembly and personal computer assembly) from the dataset, and randomly sample videos from three participants for each task.

\end{itemize}

\input{insert_figures/dataset}

\input{insert_figures/testbed}
After collecting videos with step annotations from these datasets, unless otherwise specified, we sample frames from the raw videos at 10 FPS, which is commonly used in video understanding tasks for procedural tasks~\cite{miech2020end,shen2021learning}.

\textbf{\textit{Sample Annotation Procedure.}}
As shown in Figure~\ref{fig:dataset_sample}, each sample the curated dataset consists of $\mathcal{S} = \{I_t, t, \mathbf{I}_{prev}, D_s, S_s, H_p, M_h, A_u, P_l, P_r\}$, where $I_t$ is the current frame, $t$ is the timestamp, $\mathbf{I}_{prev}=\{I_{t-9},\dots,I_{t-1}\}$ are the previous nine frames, and the remaining terms denote step description, execution status, historical task progress, hand motion cues, motion-aware action understanding, proactive trigger, and step-aware proactive response, respectively.
The step description $D_s$ is obtained from the original key-step annotations in public datasets.
As shown in Figure~\ref{fig:dataset_curation}, the rest are generated through a two-stage procedure that combines human expertise with LLM-assisted generation.
Detailed procedures are in Appendix~\ref{appendix:annotation}.

In total, our curated dataset\footnote{The curated dataset is available at \url{https://github.com/Columbia-ICSL/Pro2Assist}.} contains 1,089 samples covering ten common daily procedural tasks (i.e., making coffee, oatmeal, tea, pinwheels, quesadilla, cheese sandwich, hotdog, peanut butter sandwich, as well as PC assembly and tent assembly).
For supervised fine-tuning~(SFT) in experiments, we randomly split the dataset into 65\% for training and 35\% for testing.

\noindent\textbf{Real-world Evaluation.} 
While our curated dataset provides annotated egocentric visual data for proactive procedural assistance, it lacks paired multimodal sensor streams required for end-to-end system evaluation.
\revision{Therefore, we conduct real-world experiments and a user study on our real-world testbed.}
Specifically, we recruited 20 participants to perform assigned procedural tasks. 
Figure~\ref{fig:testbed} illustrates the testbed and data collection setup.
Each participant wears RayNeo X3 Pro smart glasses to capture synchronized multimodal data during the task.
The study was approved by the authors' institutional IRB, and all participants provided informed consent.
After data collection, participants review their recorded videos, segment them into temporal segments, and annotate each segment with step descriptions ($D_s$), step status ($S_s$), and proactive labels ($P_l$).
They also participated in a user study and completed a questionnaire based on their experience.
In total, we collected 20 multimodal recordings with an average duration of $325.4\,\text{s}$, covering five different procedural tasks from the curated dataset~(i.e., making tea, quesadilla, cheese sandwich, hotdog and peanut butter sandwich).

\subsubsection{Evaluation Metrics}
\label{sec:metrics}
We extensively evaluate \workname's performance from five perspectives as follows.

\noindent\textbf{Procedural Action Understanding Accuracy.}
\revision{This dimension assesses the system's ability to correctly infer \textit{which step} the user is performing and \textit{how far along} they are within that step (i.e., execution status). 
\textit{Step-Acc} (Step Identification Accuracy) measures whether the predicted step matches the ground truth, and \textit{Status-Acc} (Status Identification Accuracy) evaluates whether the fine-grained execution status (i.e., \textit{just start}, \textit{in progress}, \textit{about to finish}, and \textit{step transition}) is correctly identified.
For example, given the sample in Figure~\ref{fig:dataset_sample}, \textit{Step-Acc} evaluates whether the system correctly identifies the current step as ``Transfer water to kettle'' rather than other steps in the procedural task (e.g., ``Measure 12 ounces of water''), and \textit{Status-Acc} evaluates whether it correctly recognizes the status as \textit{about to finish} rather than other execution statuses.}

\noindent\textbf{Proactive Trigger Performance.}
This aspect evaluates whether proactive assistance is initiated at appropriate moments.
Following prior work~\cite{yang2025contextagent}, we use \textit{Acc-P} (Proactive Accuracy) to measure the correctness of proactive trigger predictions, \textit{MD} (Missed Detection) to quantify the rate of failing to trigger assistance when needed, and \textit{FD} (False Detection) to measure the rate of incorrectly triggering assistance when not needed.

\noindent\textbf{Proactive Timing Accuracy.}
The timing of step-aware proactive assistance is critical, as responses delivered too late may fail to help the user, while responses triggered at incorrect steps provide little value.
\revision{Moment-level metrics (e.g., \textit{Acc-P}, \textit{MD}, \textit{FD}) only evaluate whether the system correctly triggers proactive assistance, but do not capture \emph{how timely} the response is within the valid proactive period. 
In procedural tasks, a response triggered early in the proactive period is more useful than one triggered at the last moment. To evaluate timing quality, we introduce the \textit{Step-aware Timeliness Score (STS)}, a period-level metric ranging from 0 to 1, where a higher score indicates earlier and more useful proactive assistance.}
For a predicted trigger time $\hat{t}_i$ with predicted step $\hat{D}_{s_{i}}$, given ground-truth step $D_{s_{i}}$ and the corresponding proactive time period $[s_i, e_i]$, STS is defined as
\begin{equation}
\begin{split}
\text{STS}_i = 
\begin{cases}
\exp\left(-\frac{\hat{t}_i - s_i}{e_i - s_i}\right) & \text{if } \hat{D}_{s_{i}}=D_{s_{i}} \text{ and } s_i \leq \hat{t}_i \leq e_i \\
0 & \text{otherwise}
\end{cases}
\label{eq:sts}
\end{split}
\end{equation}
This metric assigns higher scores to earlier responses within the valid proactive period for the correct step, with scores decaying exponentially toward the end of the period.
A response at the final moment of the period (STS $\approx 0.368$ when $\hat{t}_i=e_i$) is still valued more than a missed or misaligned response (STS $=0$).
Overall, STS is computed by averaging across all proactive moments.
\revision{By considering temporal position within the valid proactive period, STS goes beyond binary trigger correctness and captures the practical utility of proactive assistance.}

\noindent\textbf{Response Quality.}
\revision{To measure the relevance and usefulness of generated proactive responses, we} adopt an LLM-as-a-Judge approach, whose effectiveness has been demonstrated in prior work~\cite{gu2024survey,chen2024mllm}.
We incorporate the ground-truth step, execution status, and task-specific guideline into an evaluation prompt, and use GPT-5~\cite{singh2025openai} to assess response quality.
Prompt details are provided in Appendix~\ref{appendix:prompts}.

\noindent\textbf{System Overhead.}
We measure system overhead from four aspects to evaluate \workname's efficiency for real-world deployment.
\textit{Inference Ratio} denotes the proportion of moments in which the VLM reasoner is invoked for step-aware proactive reasoning.
\textit{Proactive Hit Rate} measures the percentage of VLM inferences that occur within ground-truth proactive periods.
\revision{These two metrics should be interpreted together, as a lower \textit{Inference Ratio} is desirable only when \textit{Hit Rate} remains high, indicating selective yet effective trigger of VLM inference.}
\textit{Latency} includes inference latency and communication delay. 
\textit{Power Consumption} measures the average power usage of the smart glasses during system operation.

\subsubsection{Baselines}
We evaluate \workname~and compare it with strong baselines that are adapted to the proactive procedural assistance setting, 
\revision{including an existing procedural assistant extended with proactive capabilities (VLM-Procedure), few-shot prompting strategies adapted for proactive procedural reasoning (Vanilla ICL, ICL-EN, and CoT), and existing general proactive systems adapted to procedural tasks (VideoLLM-online and ProAgent).}

\noindent\textbf{VLM-Procedure}. 
PrISM-Q\&A~\cite{arakawa2024prism} is an LLM-based reactive procedural assistant originally designed to operate using audio, IMU, and task knowledge. 
We employ its system prompt and extend it to support VLM-based reasoning and visual inputs.
We further adapt the prompt to support proactive reasoning, enabling the model to perform proactive reasoning and generate proactive assistance.

\noindent\textbf{Vanilla ICL}. 
This baseline uses in-context learning (ICL)~\cite{dong2024survey} with few-shot demonstrations that contain only raw sensory context, relying on the VLM’s intrinsic knowledge to perform action understanding and proactive reasoning for multi-step procedural tasks.
\revision{It serves as a minimal baseline for the prompting-based approaches. Together with ICL-EN and CoT, these baselines evaluate whether widely adopted prompting techniques can sufficiently address proactive procedural assistance.}

\noindent\textbf{ICL-EN}.
Built upon the Vanilla ICL baseline, this approach explicitly incorporates task-specific expert knowledge into the system prompt, providing additional guidance for understanding procedural actions.
\revision{This evaluates to what extent expert knowledge injection via prompting can further improve performance.}

\noindent\textbf{CoT}. 
\revision{This approach employs a concise Chain-of-Thought~\cite{wei2022chain} strategy with few-shot examples containing explicit thought traces that demonstrate how to map visual cues to the current procedural step, its execution status, and step-aware proactive assistance.
This evaluates to what extent explicit structured reasoning via prompting can further improve performance.}

\noindent\textbf{VideoLLM-online}. 
VideoLLM-online~\cite{chen2024videollm} is an online VLM designed for streaming video, which introduces a Streaming EOS (End-of-Sequence) prediction objective at the model level to enable proactive response updates.
We use the released \texttt{VideoLLM-online-8B-v1+} model as a baseline in our real-world evaluation, and primarily compare it with \workname~on proactive prediction performance.
\revision{Together with ProAgent, these baselines evaluate whether existing proactive methods can generalize to long-horizon procedural tasks.}

\noindent\textbf{ProAgent (Vanilla\&FT)}. 
ProAgent~\cite{yang2025proagent} is a proactive assistance system designed for general daily scenarios based on holistic scene understanding. 
For a comprehensive comparison, we evaluate ProAgent under two configurations. 
ProAgent (Vanilla) uses the original model trained on the CAB-Lite dataset~\cite{yang2025contextagent}, following its original settings.
ProAgent (FT) is further fine-tuned on our curated dataset to better adapt it to procedural tasks.
Since ProAgent is not explicitly designed for procedural action understanding, we primarily compare it with \workname~on proactive prediction performance.

For baselines without task-specific expert knowledge, we provide the complete set of possible steps in the curated dataset to ensure fair step identification. 
For few-shot demonstrations, we randomly include five examples from the dataset in the prompt.
In the real-world evaluation, we implement VLM-Procedure with periodic sampling at $0.5\,\text{s}$ intervals to enable continuous perception and proactive reasoning.
For the ICL, ICL-EN, and CoT baselines, visual data are sampled at 10 FPS and processed by Reducto~\cite{li2020reducto} to remove redundant frames before VLM inference, improving efficiency in real-world settings.
For ProAgent, we adapt its on-demand tiered perception to procedural tasks by setting the low-rate and high-rate sampling intervals to $1\,\text{s}$ and $0.5\,\text{s}$, respectively.

\subsection{Overall Performance}
\subsubsection{Quantitative Results}
\label{sec:quantitative_results}
In this section, we evaluate the overall performance of \workname~on both the real-world testbed and the curated dataset.

\noindent\textbf{On real-world testbed.}
\input{insert_figures/overall_performance_realworld}
As shown in Figure \ref{fig:overall_realworld}, we compare \workname~with multiple baselines on the real-world evaluation.
While VideoLLM-online achieves low inference latency and 68.4\% \textit{Acc-P}, it performs poorly on procedural action understanding, with only 32.6\% \textit{Step-Acc}, which further degrades proactive timing accuracy to 33.0\% \textit{STS}.
This is because VideoLLM-online is specifically designed to efficiently process dense video streams, but does not model user intent and procedural knowledge that are essential for proactive assistance in procedural tasks.
VLM-Procedure achieves better procedural action understanding with expert knowledge, but performs poorly in proactive reasoning when relying on the model’s intrinsic knowledge to provide proactive assistance, \revision{indicating that prompt-level extension of a reactive procedural assistant cannot acquire effective proactive capability.}
\revision{Prompting techniques produce partial improvements with limited overall gains.}
\revision{Specifically, among the prompting baselines, ICL-EN and CoT exhibit distinct improvement patterns over Vanilla ICL.
Specifically, ICL-EN primarily improves procedural action understanding (4.4\% \textit{Step-Acc}, 5.1\% \textit{Status-Acc}) but decreases \textit{Acc-P}, while CoT primarily improves proactive prediction (3.8\% \textit{Acc-P}) without improving procedural action understanding,
demonstrating that incorporating expert knowledge and reasoning traces via prompting benefits different parts of the proactive procedural task.
However, the improvements among prompting baselines remain confined to 3--5\%, and the overall performance of all three remains limited, indicating that widely adopted prompting techniques alone are insufficient for proactive procedural assistance.
}
ProAgent (Vanilla), which is primarily designed for general daily scenarios, exhibits limited performance on \revision{long-horizon} procedural tasks, \revision{as it lacks explicit designs for procedural knowledge modeling and capturing temporal context in procedural tasks.}
Even after fine-tuning on the curated dataset, ProAgent (FT) still underperforms \workname, as it lacks explicit designs for procedural knowledge and temporal context, both of which are critical for procedural tasks.
Since ProAgent cannot reliably identify the current procedural step, we relax the step-matching condition in the \textit{STS} computation for ProAgent.
However, ProAgent still achieves a lower \textit{STS} than \workname.
Overall, compared to the best-performing baselines, \workname~still obtains improvements of 25.2\% in \textit{Step-Acc}, 21.6\% in \textit{Status-Acc}, and 15.1\% in \textit{Acc-P}.
Moreover, it achieves up to 2.29$\times$ the STS of the baselines.
\workname~maintains an inference latency within $0.5\,\text{s}$, slightly higher than baselines such as ICL but with a substantially lower VLM inference ratio, demonstrating an effective trade-off between efficiency and performance.
\revision{Together, these results show that no existing approach captures all the capabilities required for proactive procedural assistance, whereas \workname~addresses them jointly through its integrated design, validating its effectiveness for real-world procedural task assistance.}

\noindent\textbf{On the curated dataset.}
\input{insert_figures/overall_performance_benchmark}
As shown in Figure~\ref{fig:overall_benchmark}, we further evaluate \workname~and baseline methods on the curated dataset, focusing on the overall capability of the VLM reasonser for procedural action understanding and step-aware proactive reasoning.
Overall, \workname~significantly outperforms all baselines across all evaluation metrics.
In particular, \workname~achieves 93.6\% \textit{Step-Acc} and 77.2\% \textit{Status-Acc}, indicating effective procedural action understanding.
Besides, It achieves 86.9\% \textit{Acc-P} with \textit{MD} and \textit{FD} both below 8\%, demonstrating its ability to accurately identify moments requiring proactive assistance while avoiding unnecessary or premature interventions.
Moreover, \workname~achieves the highest scores in both \textit{Reference} and \textit{Usefulness}, indicating its assistance is not only timely but also contextually appropriate and useful.

\subsubsection{Qualitative Results}
\input{insert_figures/exp_process_bar}

As shown in Figure~\ref{fig:attention_performance}, compared with baselines, \workname~not only triggers VLM inference at appropriate moments with high proactive demand, but also more accurately identifies proactive moments to deliver timely assistance.
In contrast, baselines either trigger inference excessively, resulting in redundant predictions, or fail to reliably identify proactive moments, leading to high missed detection and false detection rates.
Figure~\ref{fig:overall_example} further presents representative examples of \workname’s inference results on a task recording, demonstrating that \workname~explicitly reasons over sensory and procedural contexts to achieve reliable action understanding.
When no assistance is needed, \workname~remains silent to avoid interrupting the user. Otherwise, it generates step-aware assistance based on the user’s current state and expert knowledge to effectively help the user perform tasks.

\input{insert_figures/overall_performance_example}

\subsection{Effectiveness of System Module}
\subsubsection{Impact of Motion-based Perception}

We evaluate \workname's \revision{motion-based perception strategy} from two perspectives, including system overhead and overall prediction performance.
\revision{As shown in Figure \ref{fig:inference_hit}, we first evaluate the impact of this strategy on system overhead by removing head motion–aware sampling (``w/o Sampling''), motion-based key moment selection (``w/o Selection''), and both components (``w/o Both'').}
\revision{
The ``w/o Selection'' and ``w/o Both'' variants have high inference ratios with low hit rates, indicating frequent but ineffective VLM inference.
The ``w/o Sampling'' variant yields a lower \textit{Inference Ratio} but also a lower \textit{Hit Rate}, as uniform-interval sampling fails to capture key moments indicated by head motion that require proactive assistance,
\final{which is further supported by the lower values of both metrics for ``w/o Both'' compared to ``w/o Selection''.}
In contrast, \workname~achieves the best trade-off with the highest \textit{Hit Rate} and a slightly higher \textit{Inference Ratio}, indicating its effectiveness.}
Furthermore, we evaluate the impact of motion extraction in motion-based key moment selection on overall prediction performance.
As shown in Figure~\ref{fig:ablation_realworld}, compared with the variant without motion extraction \revision{(``w/o Motion Extraction'')}, \workname~achieves improvements \revision{across metrics, indicating} that incorporating motion extraction allows \workname~to leverage fine-grained hand motion cues for more accurate procedural action understanding, which in turn enhances step-aware proactive prediction.

\input{insert_figures/exp_attention}

\subsubsection{Impact of Motion-Aware Action Understanding}
As shown in Figure~\ref{fig:ablation_study}, removing the motion-aware action understanding objective during SFT \revision{(``w/o Action Understanding'')} degrades \workname's performance on the curated dataset by 2.1\% in \textit{Step-Acc}, 6.8\% in \textit{Status-Acc}, and 6.2\% in \textit{Acc-P}, demonstrating its effectiveness for procedural proactive assistance.

\subsubsection{Impact of Temporal Context and Expert Knowledge}
We examine variants that remove temporal context, expert knowledge, and both components \revision{(denoted as ``w/o Temporal Context'', ``w/o Expert Knowledge'', and ``w/o EK+TC'', respectively)}.
As shown in Figure~\ref{fig:ablation_study}, all the variants lead to significant performance degradation. 
For example, \workname~achieves improvements of 19.8\% in \textit{Step-Acc}, 
3.4\% in \textit{Status-Acc}, and 5.4\% in \textit{Acc-P} over the 
\revision{``w/o EK+TC''} variant.
Overall, the results validate the effectiveness of incorporating them into reasoning.

\input{insert_figures/exp_ablation}

\subsubsection{Impact of Step-Aware Consistency Checking}
We evaluate the mechanism by removing it \revision{(``w/o Consistency Checking'')} in the real-world evaluation. 
First, we evaluate its effectiveness in reducing unnecessary user interruptions.
As shown in Figure~\ref{fig:display_checker}, with the mechanism, \workname~delivers new assistance only when the user's state changes, avoiding repetitive guidance and reducing perceived intrusiveness by over 50\%.
Second, we evaluate its effectiveness in preventing single-moment mispredictions from degrading overall performance.
As shown in Figure~\ref{fig:ablation_realworld}, removing it results in a drop of over 20\% in both \textit{Step-Acc} and \textit{STS}, indicating that single-moment mispredictions accumulate and degrade subsequent reasoning.
These results demonstrate that the mechanism is crucial for both user experience and robust reasoning.

\subsubsection{Impact of Hyper-parameters}
We further evaluate the effects of parameter settings.

\noindent\textbf{Parameter Sensitivity in \revision{Motion-Based} Perception}
We analyze the impact of the sampling and filtering thresholds on \textit{Inference Ratio} and \textit{Proactive Hit Rate}, which jointly evaluate the tradeoff between computational efficiency and sampling effectiveness.
As shown in Figure~\ref{fig:thr_setting}, we sweep one threshold while fixing the other.
For motion-aware sampling, a low threshold cannot help with avoiding unnecessary VLM inference, while an overly large threshold makes the system miss attention shifts indicated by head motion, reducing hit rate.
Similarly, a low filtering threshold keeps many frames with minimal motion, while an excessively high threshold may filter out frames requiring assistance, negatively affecting proactive performance.
\final{Within a reasonable range of threshold values, both metrics remain stable, reflecting the robustness of the perception mechanism, which effectively combines low-cost, always-on head motion signals with fine-grained hand motion cues.}
Overall, across a wide range of settings, \workname~consistently outperforms VLM-Procedure with periodic sampling (20\% inference ratio, 27\% hit rate), indicating that \revision{motion-based} perception provides a better efficiency–effectiveness tradeoff.

\input{insert_figures/exp_base_model}

\noindent\textbf{Impact of Base VLM Models.}
As shown in Figure~\ref{fig:base_model}, we further evaluate \workname~with different VLM models as the base model.
\revision{The results demonstrate that \workname~works effectively across different base VLMs, and scaling up the base model consistently improves its performance.}

\noindent\textbf{Impact of Temporal Window Length in Consistency Checking.}
We vary the window length in the real-world evaluation.
As shown in Figure~\ref{fig:checker_window_len}, across all settings, \workname~consistently and significantly outperforms the variant without this checking mechanism~(denoted as ``w/o Checking'')\revision{, indicating the effectiveness and robustness of \workname~with respect to the choice of consistency history length.}

\subsection{Out-of-Domain Evaluation}
\revision{We evaluate \workname’s ability to generalize to unseen tasks} by randomly splitting the curated dataset at the procedural task level. Samples from six tasks are used for training, while the remaining four tasks are reserved for evaluation.
As shown in Figure~\ref{fig:ood_performance}, \workname~remains effective across different base VLMs on unseen tasks. 
We further assess \workname~on a subset of the real-world dataset that includes three procedural tasks unseen during training (i.e., making tea, quesadilla, and cheese sandwich)\revision{, where} it achieves on average 82.2\% \textit{Step-Acc}, 71.0\% \textit{Status-Acc}, 75.3\% \textit{Acc-P}, and 67.2\% \textit{STS}\revision{, indicating its generalization in real-world settings.}

\subsection{System Overhead}
\label{sec:system_overhead}

\input{insert_figures/exp_overhead}

We evaluate the system overhead of \workname~across multiple back-end platforms.
\final{The total inference time consists of VLM inference, hand detection, and motion extraction.}
As shown in Figure~\ref{fig:latency}, \workname~achieves a total inference time of $0.49\,\text{s}$ on an NVIDIA RTX 5090 GPU and $4.51\,\text{s}$ on an NVIDIA Jetson Orin.
Notably, the time to first token (TTFT) consistently remains below $330\,\text{ms}$, enabling \workname~to begin delivering assistance promptly via streaming output \final{for real-time responsiveness}, even on resource-constrained edge devices.
Moreover, expert knowledge retrieval takes under $0.22\,\text{s}$ across all devices, and the average communication latency in two real-world evaluation environments is $327.3\,\text{ms}$.
In addition, we measure the average power consumption of the smart glasses while running \workname, which is $2.2\,\text{W}$, indicating it is practical for real-world deployment.

\subsection{User Study}
We conduct a user study with 20 participants (10 male and 10 female\revision{, P1-P20}) with an average age of 27, whose education levels ranged from undergraduate to Ph.D., to evaluate whether \workname~meets expectations as a proactive procedural assistant.
Participants provided feedback for \workname~'s assistance through a questionnaire containing three parts with nine questions in total.
\revision{Following prior studies\cite{yang2025socialmind,mccloud2022using,emami2021privacy}, we applied categorical ratings with distributional analysis for characterizing users' perception of the system, with questions and question-specific response options phrased in plain language, so that participants can interpret each option directly regardless of technical background.}
\revision{The questionnaire contains three parts that capture participants' background, subjective system evaluation, and preferences for proactive procedural assistants, with the specific questions chosen based on \workname's design as an AR glasses-based assistant and key dimensions adopted by prior proactive and procedural assistance studies~\cite{pu2025promemassist,yang2025socialmind,huang2025vinci}.}
Details are as follows.
\begin{itemize}[leftmargin=*]
\item 
\textbf{{S1. Background Information.}}
This part collects participants’ prior experience with the task and frequency of performing it, as well as their previous use of smart assistants in procedural tasks.

\item 
\textbf{{S2. System Evaluation.}}
\textit{Contextual Relevance} assesses whether the delivered messages match the participant’s current step and execution status.
\textit{Timeliness} assesses whether messages are delivered at appropriate moments.
\textit{Usefulness} assesses whether messages help participants complete the task.
\textit{Intrusiveness} assesses whether the system is disruptive due to excessive messages.
\textit{Willingness} assesses participants’ willingness to use the system in the future.

\item 
\textbf{{S3. System Preferences.}}
This part examines user preferences for proactive assistant design, including acceptable response latency and preferred delivery method for proactive assistance.
\end{itemize}

\input{insert_figures/user_study}
\noindent\textbf{Overall User Perception of \workname}. Figure~\ref{fig:user_study} demonstrates participants' feedback.
Among the participants, 60\% had no prior experience with the assigned task, and only two had previously used smart assistants for procedural activities.
Overall, 90\% participants found the system useful, with particularly strong agreement among those without prior task experience.
This indicates that \workname~is especially beneficial for users learning new procedural tasks, as it continuously provides guidance aligned with the task workflow.
Regarding contextual relevance, 75\% reported that the assistance closely matched their progress and ongoing actions.
For timeliness, 40\% rated the system as excellent and 55\% as acceptable, indicating that the assistance is generally delivered at appropriate moments.
Regarding intrusiveness, 55\% found the system as minimally intrusive, while 20\% felt that the assistance was somewhat lengthy or excessive. Notably, most of these 20\% had prior task experience and agreed that the proactive timing and content were appropriate, but preferred more concise responses. 
For system preferences, 35\% desired assistance delivery within $1\,\text{s}$, while the majority found longer latencies acceptable.
\workname~meets these expectations in most scenarios, achieving an average end-to-end latency within $5\,\text{s}$ on an edge device and can be reduced to within $1\,\text{s}$ on GPU platforms (details are in \S\ref{sec:system_overhead}).
For delivery modality, 60\% preferred visual text overlays on the glasses display, consistent with \workname’s current design.
Another 30\% preferred audio notifications, and 10\% suggested adapting the delivery method to different task contexts.
Note that while \workname~primarily presents assistance through on-screen display, audio delivery is also supported through the smart glasses' built-in audio system.

\revision{\noindent\textbf{Analysis of Negative Experiences.}
We further analyze negative experiences, which mainly arise from four aspects as follows.}
\revision{\begin{itemize}[leftmargin=*]
\item \textit{Intrusiveness for experienced users.} 
While most participants did not find the responses intrusive, a few participants with prior task experience (P6, P13) preferred concise responses over detailed instructions. P6 mentioned, \textit{``Because I have done this task before, I just want the system to remind me of each step with concise guidance\dots it still feels intrusive to read through each time.''}

\item \textit{Unnecessary interruptions from false detections.} 
These typically occur in the middle of a step when the system is expected to remain silent.
Some participants (P2, P5, P10) reported they noticed such interruptions but considered them acceptable, as they were mainly early reminders rather than unrelated guidance.
For example, P10 mentioned, \textit{``I was still scooping peanut butter and wanted more, but it said the step was done and to move on.''} 

\item \textit{Mistimed assistance.} While 95\% of participants rated \textit{Timeliness} as excellent or acceptable, assistance can feel mistimed on short, familiar steps where users can move quickly from initiation to execution. P6, who had prior experience with the task, mentioned, \textit{``I felt it was not useful when I needed to wait for the instruction on a short and simple step.''}
This suggests timing tolerance depends on expertise, as experienced users can initiate actions quickly and have less tolerance for instructions that arrive after they are ready to execute.

\item \textit{Incorrect step guidance.} 
Such negative experiences arise from single-moment mispredictions, which are typically corrected by \workname's subsequent predictions.
While these errors might cause confusion, they were brief and resolvable in our study.
This is reflected in \textit{Relevance} ratings, where 90\% of participants reported high or moderate alignment.
Participants were able to handle such brief errors by relying on their own step awareness, informed by \workname's prior correct step-aware guidance.
P13 described, \textit{``It first regarded my reach for cinnamon as reaching for banana slices next to it, but it corrected itself once I grabbed cinnamon.''} 
This highlights that quick recovery from mispredictions is critical to sustaining user experience.

\end{itemize}}

\revision{Together with overall user perception, these findings show that \workname~delivers contextually useful and timely assistance for most participants, while revealing design implications and opportunities for further improvement, which we discuss in \S\ref{sec:discussion}.}

\input{insert_figures/user_study_baseline}
\revision{\noindent\textbf{User Experience Compared against Baselines.}}
\revision{To further compare \workname~with baselines from a user-experience perspective, we conducted a complementary study with the same 20 participants, who reviewed videos showing the proactive assistance generated by \workname~and three representative baselines (VLM-Procedure, CoT, and ProAgent~(FT)) and rated each system on a 7-point Likert scale across the five evaluation dimensions.
For each participant, all four methods were applied to the same video to control for content effects, and the resulting outputs were presented in randomized order with method identities hidden to prevent bias.
As shown in Figure~\ref{fig:user_study_baselines}, \workname~consistently outperforms all three baselines across the five dimensions, with all gains being significant ($\textit{p} < 0.001$).
These subjective gains are consistent with the results in \S\ref{sec:quantitative_results}, indicating that \workname~provides assistance that is more contextually aligned, better timed, more useful, and less intrusive, leading to higher willingness to use it.} 

%% file: insert_figures/dataset.tex
\begin{figure}[t]
    \centering
    \captionsetup{skip=3pt}
    \includegraphics[width=0.85\textwidth]{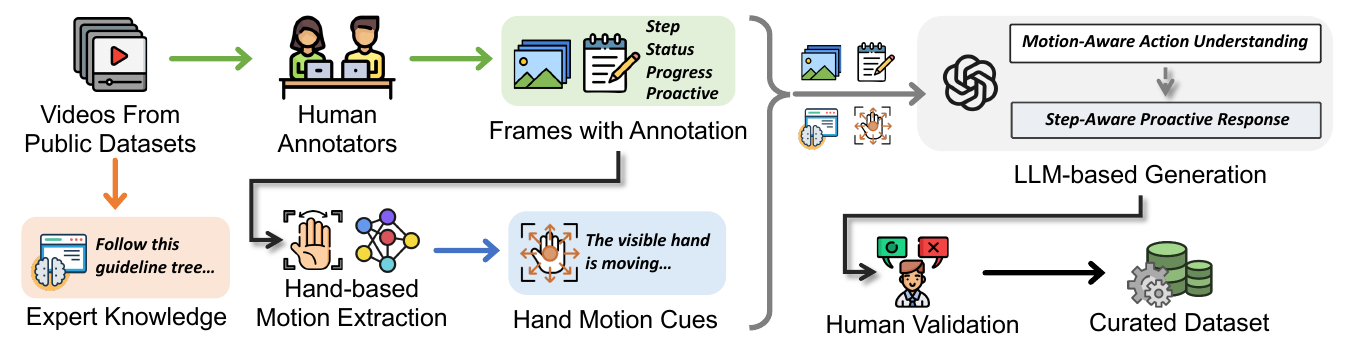}
\caption{Our dataset curation pipeline combining human annotation and LLM-assisted generation.}
    \vspace{-1em}
    \label{fig:dataset_curation}
\end{figure}

%% file: insert_figures/testbed.tex
\begin{figure*}[t]
\small
\centering

\begin{minipage}[t]{0.39\textwidth} %
\centering
\includegraphics[width=\textwidth]{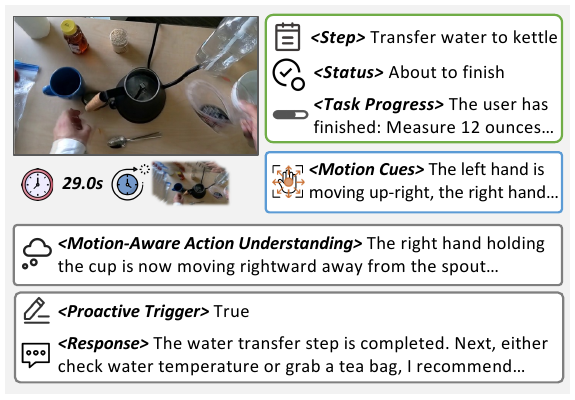}
\captionsetup{skip=5pt}
    \caption{A sample from the curated dataset, consisting of an egocentric frame and its corresponding annotations.}
    \label{fig:dataset_sample}
\end{minipage}
\hfill
\begin{minipage}[t]{0.58\textwidth} %
\centering
\includegraphics[width=0.82\textwidth]{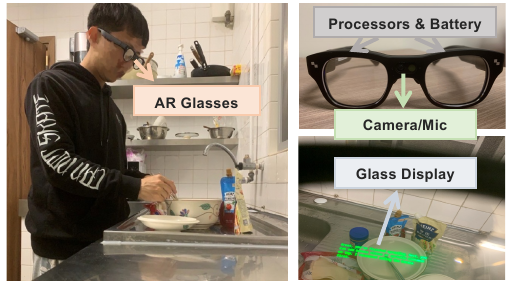}
\captionsetup{skip=5pt}
    \caption{Illustration of the real-world evaluation testbed. Participants wear AR glasses while performing procedural tasks, and \workname~displays step-aware proactive messages directly on the glasses.}
    \label{fig:testbed}
\end{minipage}%

\end{figure*}

%% file: insert_figures/overall_performance_realworld.tex
\begin{figure}[t]
    \centering
    \includegraphics[width=\textwidth]{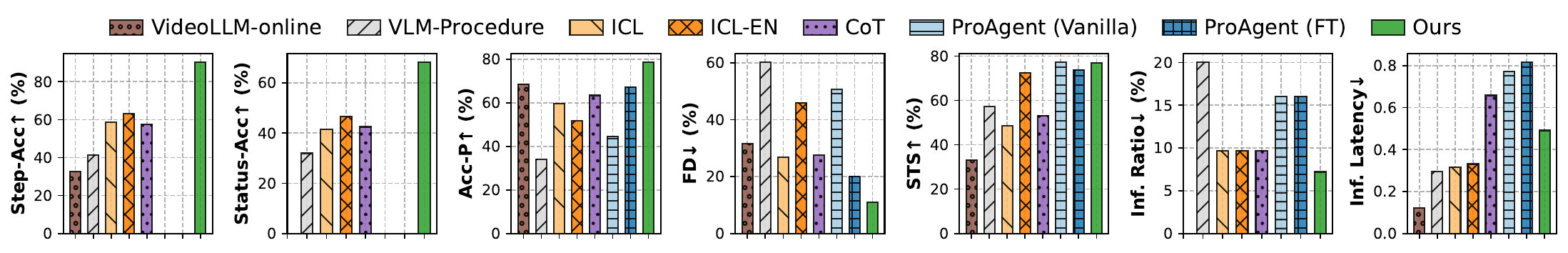}
            \vspace{-2.0em}
    \caption{End-to-end performance comparison in real-world evaluation. Missing bars indicate that the corresponding metric is not applicable to that method due to its design.
    }
    \label{fig:overall_realworld}
\end{figure}

%% file: insert_figures/overall_performance_benchmark.tex
\begin{figure}[t]
    \centering
    \includegraphics[width=\textwidth]{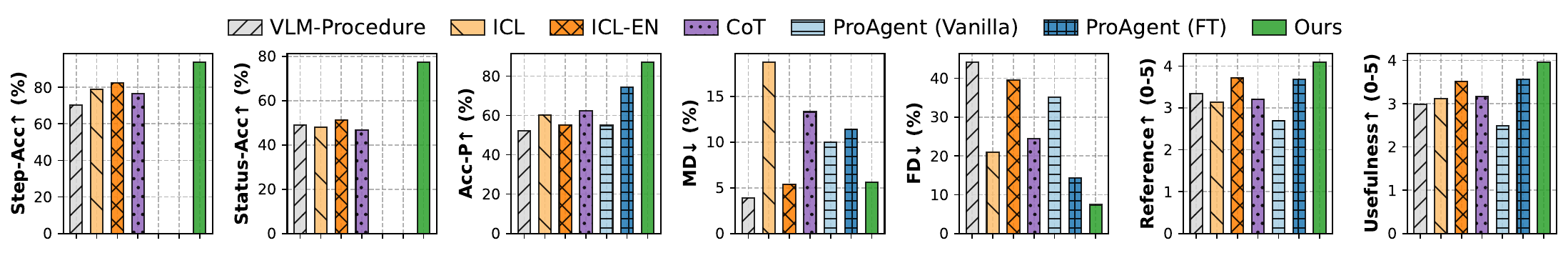}
    \vspace{-2.0em}
    \caption{Overall performance comparison on the curated dataset. 
    Missing bars indicate that the corresponding metric is not applicable to that method due to its design.}
    \vspace{-1em}
    \label{fig:overall_benchmark}
\end{figure}

%% file: insert_figures/exp_process_bar.tex
\begin{figure}[t] %
\centering
\includegraphics[width=\textwidth]{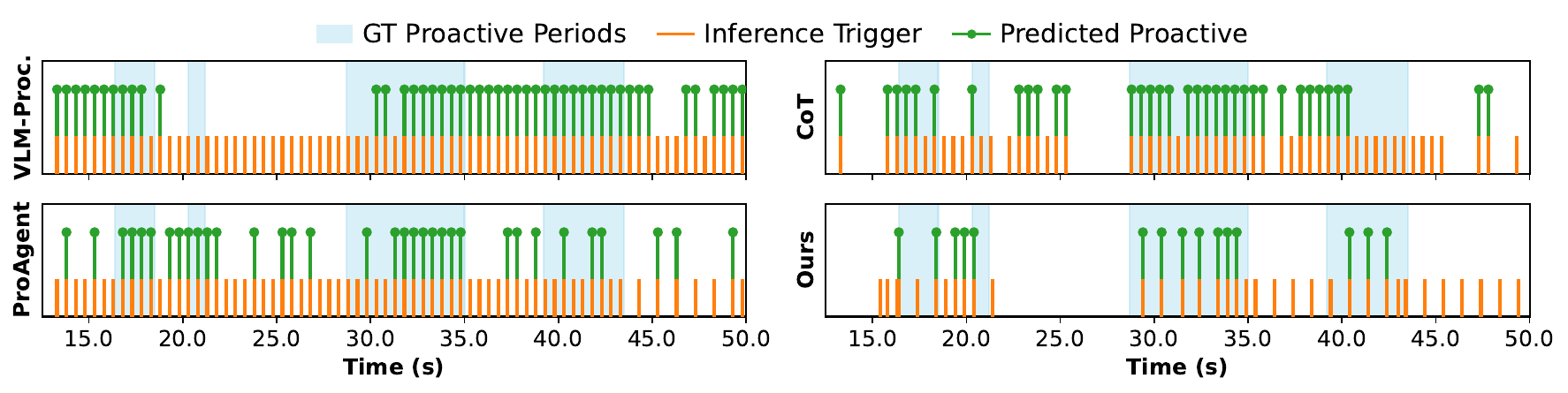}
\captionsetup{skip=1pt}
\caption{Performance comparison of inference triggering and proactive prediction. Solid orange lines indicate VLM inference timestamps, blue shaded regions denote ground-truth proactive intervals, and green lines with circular markers represent predicted proactive triggers. ``VLM-Proc.'' represents the VLM-Procedure baseline.}
\label{fig:attention_performance}
\end{figure}

%% file: insert_figures/overall_performance_example.tex
\begin{figure}[t]
    \centering
    \captionsetup{skip=5pt}
        \includegraphics[width=\textwidth]{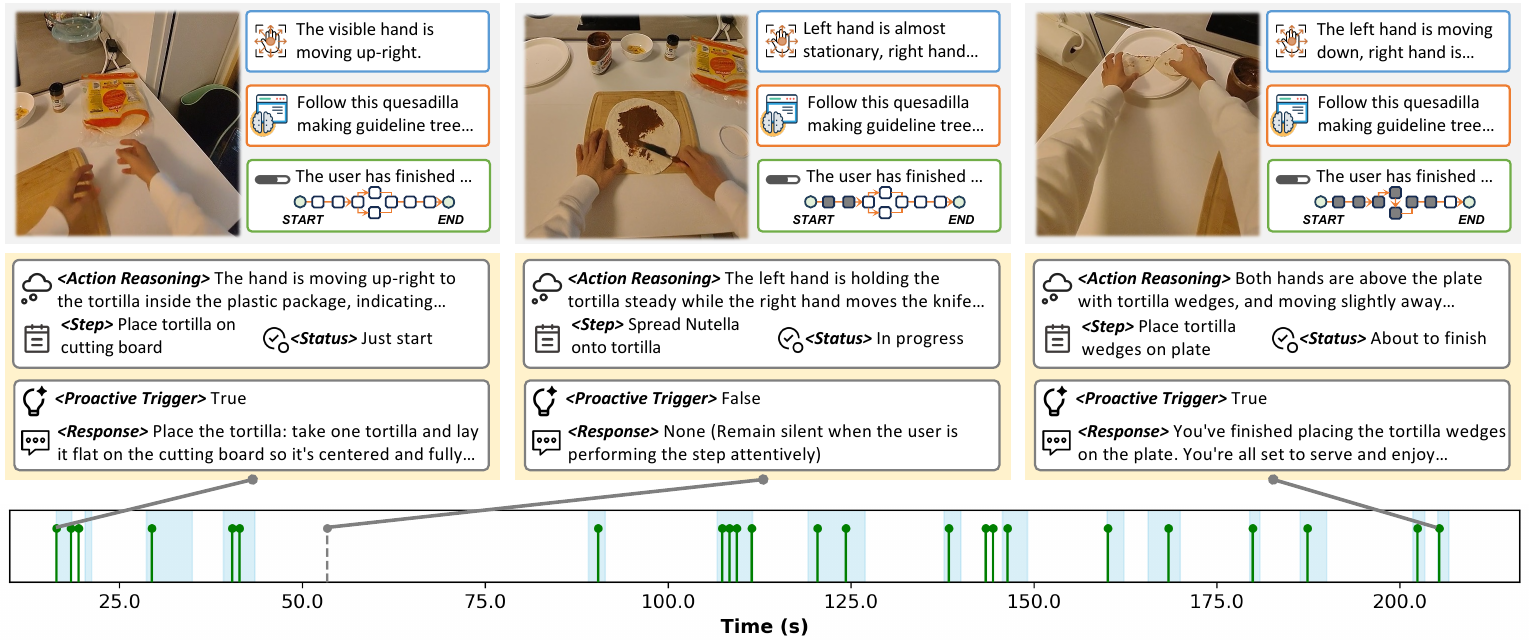}
    \caption{Examples of \workname’s inference results in real-world evaluation.}
    \vspace{-1em}
    \label{fig:overall_example}
\end{figure}

%% file: insert_figures/exp_attention.tex
\begin{figure*}[t]
\small
\centering
\begin{minipage}[t]{0.35\textwidth} %
\centering
\includegraphics[width=\textwidth]{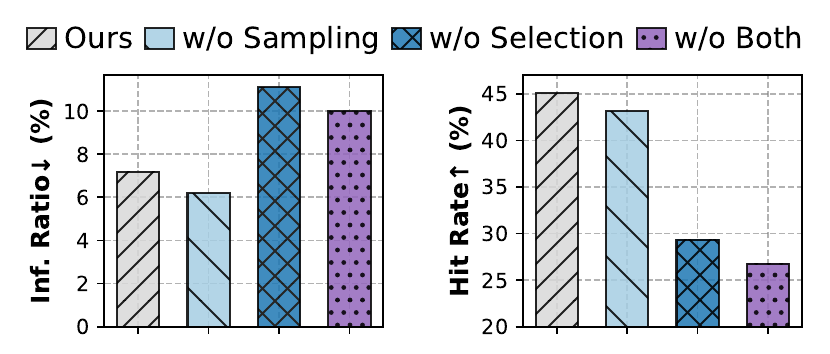}
\captionsetup{skip=1pt}
\caption{
\revision{Impact of motion-based perception on inference ratio and proactive hit rate.}
}
\label{fig:inference_hit}
\end{minipage}%
\hfill
\begin{minipage}[t]{0.64\textwidth} %
\centering
\includegraphics[width=\textwidth]{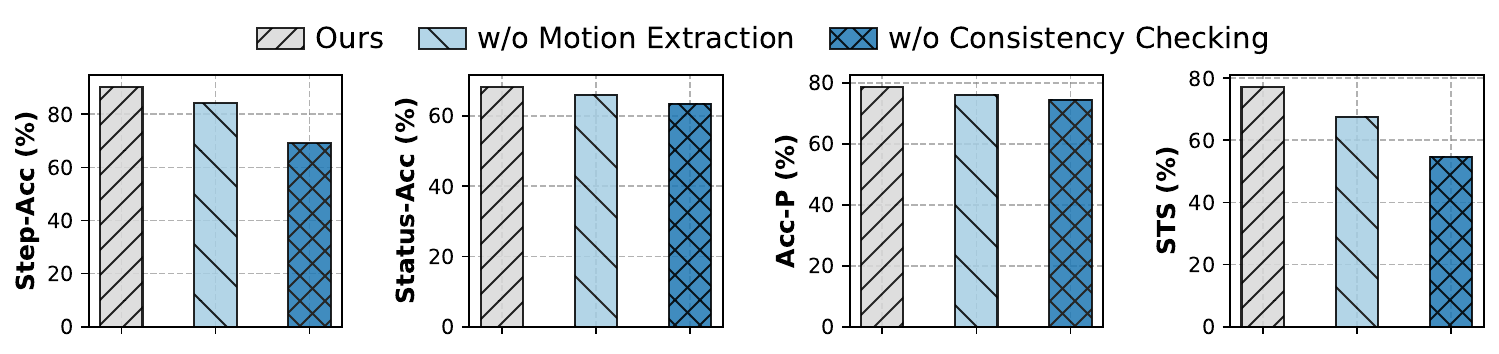}
\captionsetup{skip=1pt}
\caption{
\revision{Impact of motion extraction in motion-based perception and the step-aware consistency checking mechanism.}
}
\label{fig:ablation_realworld}
\end{minipage}

\end{figure*}

%% file: insert_figures/exp_ablation.tex
\begin{figure*}[t]
\small
\centering
\begin{minipage}[t]{0.49\textwidth} %
\centering
\captionsetup{skip=1pt}
\includegraphics[width=\textwidth]{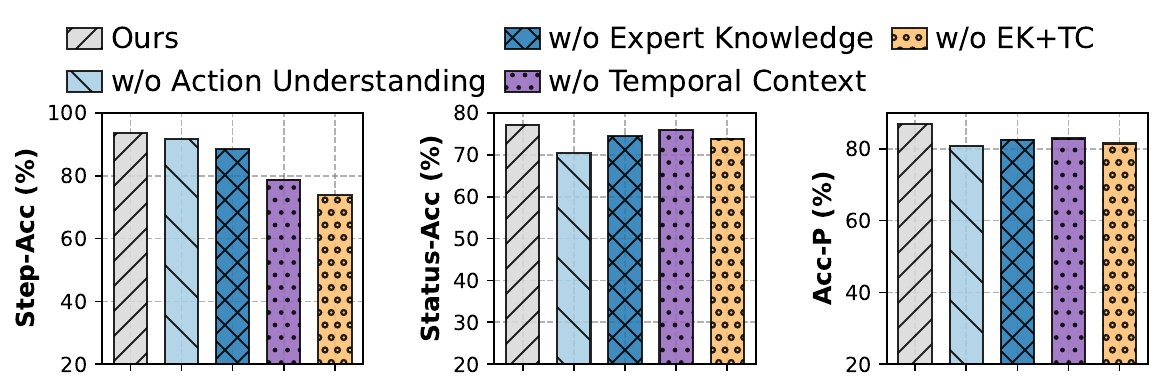}
    \caption{\revision{Ablation study of \workname's VLM reasoner. ``EK'' and ``TC'' denote expert knowledge and temporal context.}}
    \label{fig:ablation_study}
\end{minipage}%
\hfill
\begin{minipage}[t]{0.49\textwidth} %
\centering
\captionsetup{skip=1pt}
\includegraphics[width=\textwidth]{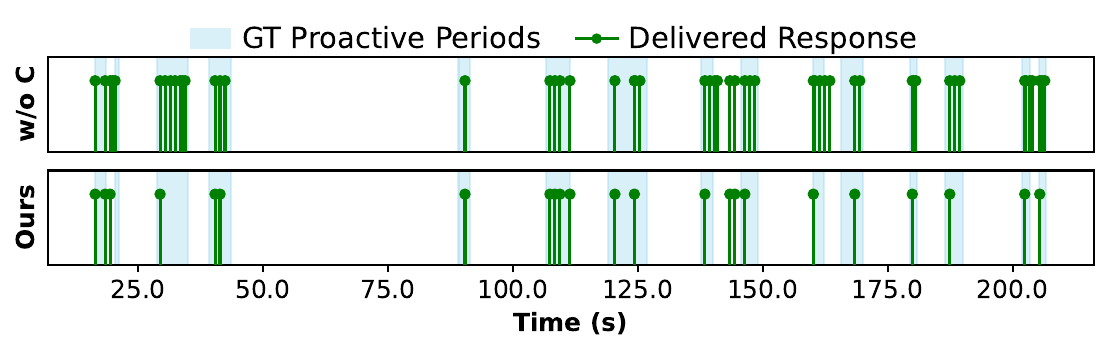}
    \caption{Impact of step-aware consistency checking on avoiding repeatedly delivering similar assistance.}
    \label{fig:display_checker}
\end{minipage}

\vspace{-1em}
\end{figure*}

%% file: insert_figures/exp_base_model.tex
\begin{figure*}[t]
\small
\centering
\begin{minipage}[t]{0.49\textwidth} %
\centering
\includegraphics[width=\textwidth]{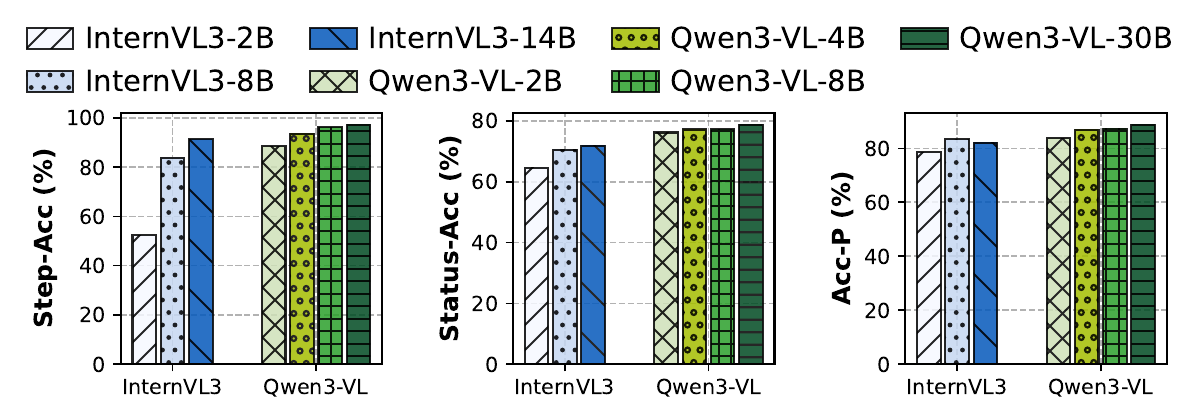}
\captionsetup{skip=1pt}
\caption{Comparison of different base VLMs used in \workname~on the curated dataset.}
\label{fig:base_model}
\end{minipage}%
\hfill
\begin{minipage}[t]{0.49\textwidth} %
\centering
\includegraphics[width=\textwidth]{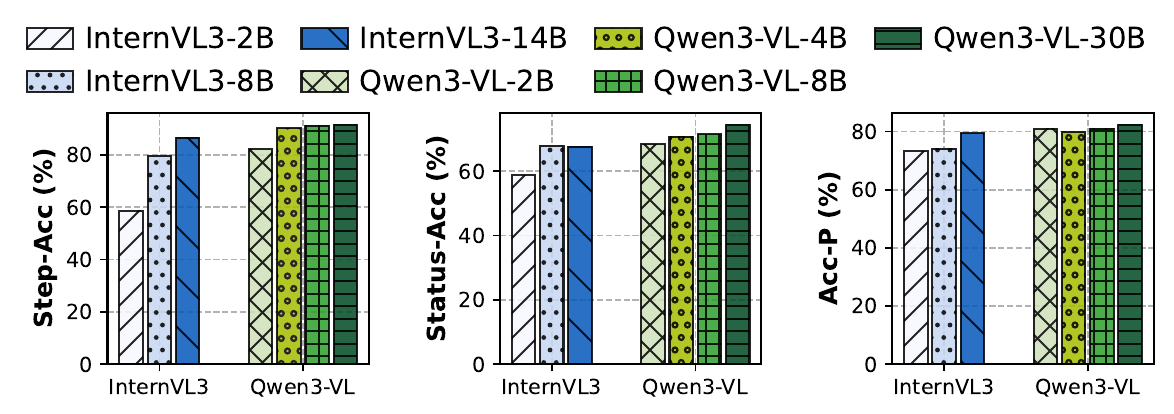}
\captionsetup{skip=1pt}
\caption{Performance of \workname~on the curated dataset in out-of-domain settings.}
\label{fig:ood_performance}
\end{minipage}
\end{figure*}

%% file: insert_figures/exp_overhead.tex
\begin{figure*}[t]
\small
\centering
\begin{minipage}[t]{0.32\textwidth} %
\centering
\includegraphics[width=\textwidth]{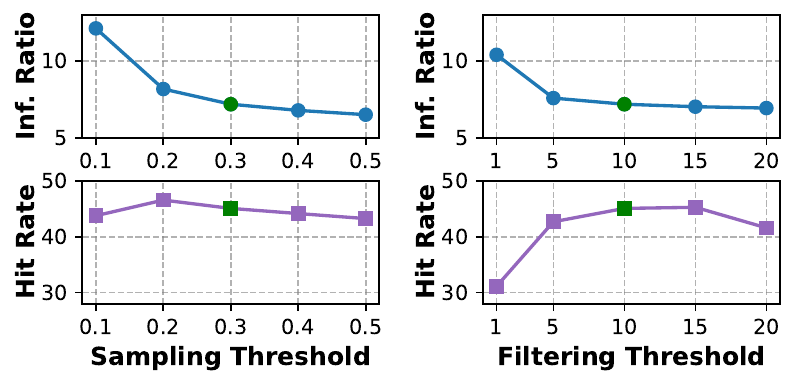}
\captionsetup{skip=1pt}
\caption{Impact of threshold settings in \revision{motion-based perception}.}
\label{fig:thr_setting}
\end{minipage}%
\hfill
\begin{minipage}[t]{0.32\textwidth} %
\centering
\includegraphics[width=\textwidth]{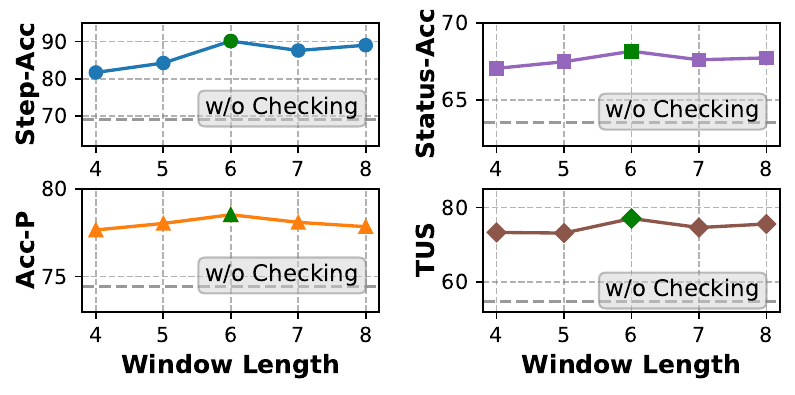}
\captionsetup{skip=1pt}
\caption{\revision{Impact of window length in step-aware consistency checking.}}
    \label{fig:checker_window_len}
\end{minipage}
\hfill
\begin{minipage}[t]{0.33\textwidth} %
\centering
\includegraphics[width=\textwidth]{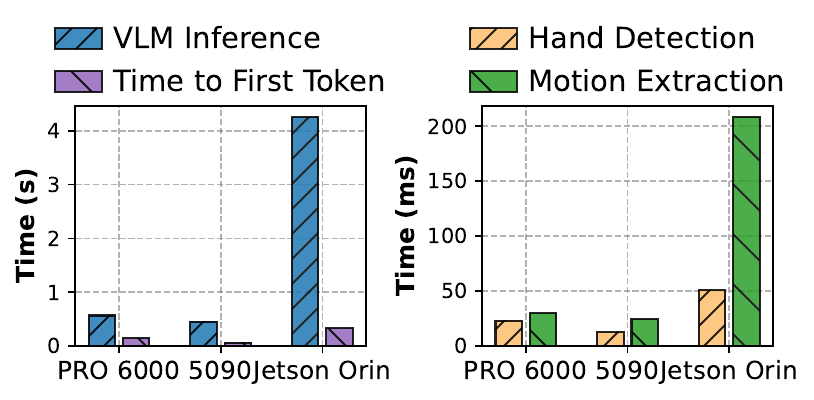}
\captionsetup{skip=1pt}
\caption{\revision{\workname's system latency across different devices.}}
    \label{fig:latency}
\end{minipage}
\end{figure*}

%% file: insert_figures/user_study.tex
\begin{figure}
    \centering
    \begin{subfigure}{0.32\columnwidth}
        \centering
        \includegraphics[width=0.86\textwidth]{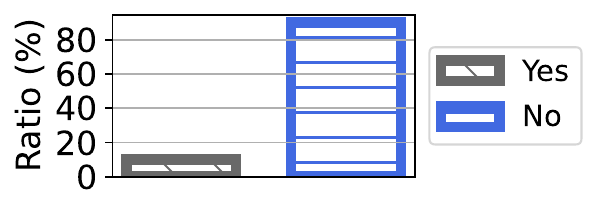}
        \vspace{-1.0em}
        \caption{Have you ever used smart assistants in procedural tasks before?}  \label{fig:user_study_assistant}
    \end{subfigure}
    \hfill
    \begin{subfigure}{0.32\columnwidth}
        \centering
        \includegraphics[width=\textwidth]{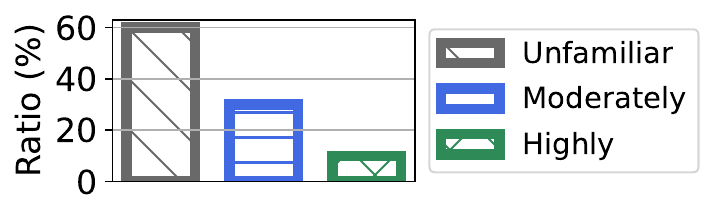}
        \vspace{-2.0em}
        \caption{How familiar are you with the procedural task?}  \label{fig:user_study_fimilarity}
    \end{subfigure}
    \hfill
    \begin{subfigure}{0.32\columnwidth}  
        \centering 
        \includegraphics[width=\textwidth]{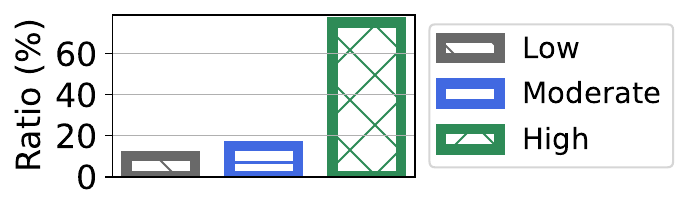}
        \vspace{-2.0em}
        \caption{How relevant was the system's assistance to what you were doing?}    
        \label{fig:user_study_relevance}
    \end{subfigure}
    \begin{subfigure}{0.32\columnwidth}  
        \centering 
        \includegraphics[width=1\textwidth]{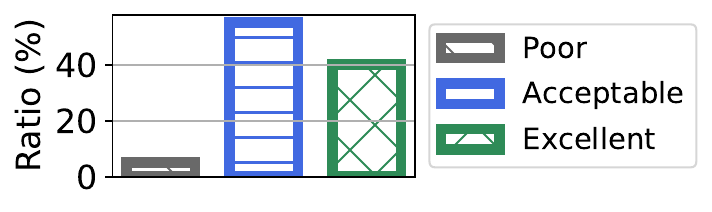}
        \vspace{-2.0em}
        \caption{How appropriate was the timing of the system's assistance?}    
        \label{fig:user_study_timeliness}
    \end{subfigure}
    \hfill
    \begin{subfigure}{0.32\columnwidth}   
        \centering 
        \includegraphics[width=1\textwidth]{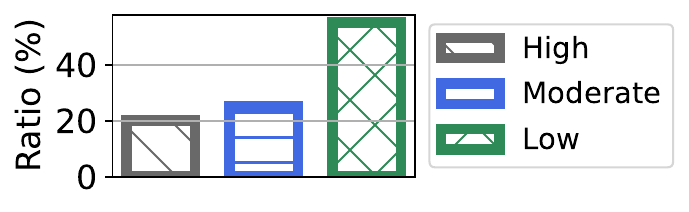}
        \vspace{-2.0em}
        \caption{How disruptive did you find the system's assistance during task execution?}    \label{fig:user_study_intrusiveness}
    \end{subfigure}
    \hfill
    \begin{subfigure}{0.32\columnwidth}   
        \centering 
        \includegraphics[width=1\textwidth]{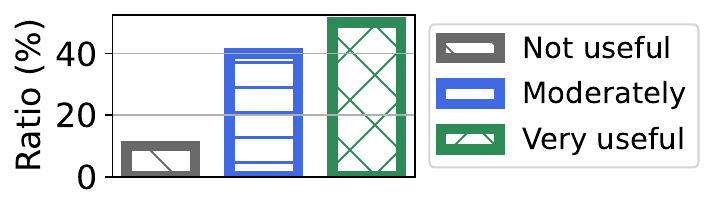}
        \vspace{-2.0em}
        \caption{How useful were the system's proactive assistance for completing the task?}    \label{fig:user_study_usefulness}
    \end{subfigure}
    \begin{subfigure}{0.32\columnwidth}  
        \centering 
    \includegraphics[width=1\textwidth]{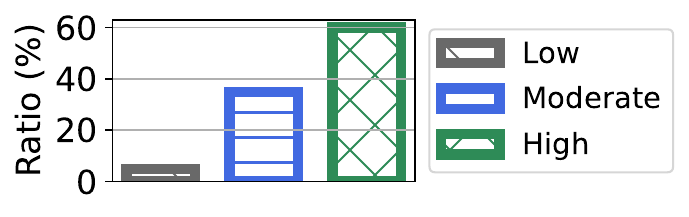}
        \vspace{-2.0em}
        \caption{How willing are you to use this system for procedural tasks in the future?}    
        \label{fig:user_study_willingness}
    \end{subfigure}
    \hfill
    \begin{subfigure}{0.32\columnwidth}   
        \centering 
        \includegraphics[width=0.9\textwidth]{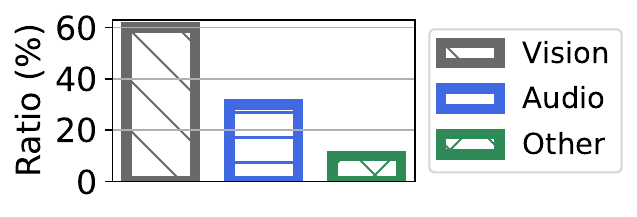}
        \vspace{-1.0em}
        \caption{What is your preferred delivery method for receiving proactive assistance during the task?}    \label{fig:user_study_deliver}
    \end{subfigure}
    \hfill
    \begin{subfigure}{0.32\columnwidth}   
        \centering 
        \includegraphics[width=1\textwidth]{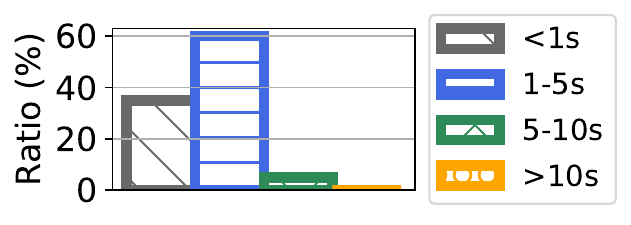}
        \vspace{-2.0em}
        \caption{What is the maximum latency you can accept for such assistance?}    \label{fig:user_study_time}
    \end{subfigure}
     \vspace{-.5em}
    \caption{\revision{Results of the overall user perception for \workname.}}
    
    \label{fig:user_study}
\end{figure}

%% file: insert_figures/user_study_baseline.tex
\begin{figure}[t]
    \centering
    \captionsetup{skip=5pt}
        \includegraphics[width=\textwidth]{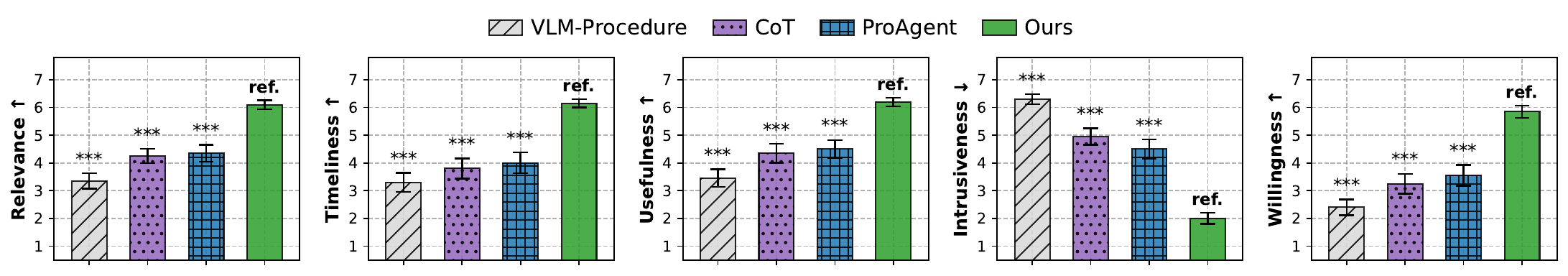}
    \caption{\revision{Comparison of baselines and \workname~using a 7-point Likert scale, ranging from 1 (Very low) to 7 (Very high). For \textit{Intrusiveness}, lower is better. ``ref.'' marks reference, and $^{***}p<0.001$ (paired Wilcoxon signed-rank test).}}
    \label{fig:user_study_baselines}
\end{figure}

%% file: section/Discussion.tex
\section{DISCUSSION AND LIMITATIONS}
\label{sec:discussion}

\revision{
In this section, we discuss the limitations and future directions of \workname, as well as design implications for proactive procedural assistants.
}

\revision{
\noindent\textbf{Sensing Modalities.}
\workname~utilizes commonly available modalities on smart glasses, among which egocentric vision and head IMU provide important cues for understanding user actions and modeling user intent.
Additional sensing modalities available on emerging smart glasses~(e.g., Meta Orion~\cite{meta2026orion}, Magic Leap 2~\cite{magicleap}) offer promising directions for future enhancement.
For example, eye gaze tracking provides direct signals of user attention~\cite{wilson2025eye} and depth sensing provides explicit geometric and spatial cues~\cite{huang2021survey,chao2021dexycb}, which could enhance finer-grained intent modeling and richer hand-object spatial understanding.
}

\revision{
\noindent\textbf{VLM Reasoning and Enhancement.}
Our evaluation across base VLMs and ablation studies shows that while stronger VLMs improve \workname's performance, our designed components remain essential for proactive procedural assistance, as they provide capabilities complementary to general VLM reasoning and enable system-level control over when to invoke inference and whether to deliver responses.
In future work, \workname~could further benefit from advances in LLM/VLM research beyond adopting stronger models.}
For instance, recent work~\cite{yang2025contextagent} has shown the effectiveness of invoking external tools to provide assistance, which could be integrated into \workname~to further enrich assistance.
Additionally, advances in efficient VLM inference~\cite{wang2023efficientvlm,yang2023edgefm,zhang2024sparsevlm} provide potential to further reduce \workname~'s latency on edge devices.

\revision{
\noindent\textbf{Error Detection in Procedural Tasks.}
Error detection is critical in procedural tasks~\cite{wang2025chef,lee2024error}.
While \workname~currently focuses on \final{continuous} step-aware guidance rather than explicit error detection, its multi-scale temporal context is promising for this extension, as hand-object interaction features~\cite{lee2024error} and comparing actions against expected steps~\cite{flaborea2024prego,wang2025chef} are effective for procedural error detection.
}

\final{
\noindent\textbf{Extension to Mixed-Initiative Assistance.}
While \workname~delivers proactive assistance to help users perform tasks smoothly without explicit queries that may interrupt ongoing operations, reactive assistance~\cite{huang2025vinci,arakawa2024prism} provides complementary benefits by giving users direct control over when and what assistance to receive through on-demand interactions.
Future work can extend \workname~toward mixed-initiative assistance that combines the advantages of both proactive and reactive assistance modes, as responses to user queries can be directly grounded in the step-aware context continuously tracked by \workname.
}

\revision{
\noindent\textbf{Extension to Longitudinal Use.}
\workname~currently focuses on continuous, single-session task execution.
Extending to procedures spanning multiple days (e.g., sourdough baking) would require identifying procedural actions from continuous daily sensing and recovering task progress upon resumption.
Additionally, leveraging cross-session experience over days for long-term personalization could enable adaptation to user-specific execution patterns for the procedural task.
\workname's multi-scale temporal context, which captures short-term hand manipulation cues and tracks per-session step progression, could be extended to support these scenarios.
Recent advances in episodic memory~\cite{wang2023lifelongmemory,luo2024video} and memory-augmented reasoning~\cite{choi2025designing} offer further enabling techniques.
}

\revision{
\noindent\textbf{Design Implications.}
Our study reveals several design implications for future proactive procedural assistants.
First, procedural assistance should minimize disruption to active task execution. Participants valued \workname's tendency to remain silent during stable execution and deliver timely guidance at step transitions, since poorly timed interventions could break the user's workflow.
Second, assistants should recover from mispredictions to sustain user experience. Participants found brief errors acceptable when quickly resolved by subsequent responses, as they can handle such errors with their own task awareness informed by prior correct guidance. 
Third, assistance may need to be adaptive through user modeling.
The study shows that the value of assistance varies with individuals, and participants less familiar with the task, who have the greatest need for guidance, benefit most from our system's continuous step-aware guidance.
Modeling user expertise and preferences would allow future systems to better adjust trigger timing and response verbosity.
}

%% file: section/Conclusion.tex
\section{CONCLUSION}
This paper introduces \workname, \revision{an end-to-end system that provides continuous, step-aware guidance during procedural tasks}.  
It leverages multimodal egocentric data from smart glasses and the reasoning capabilities of VLMs to continuously track, reason over, and assist the user's evolving task state.
Extensive evaluations show that \workname~significantly improves proactive procedural assistance, and a user study further confirms that users find it useful and are willing to adopt it in everyday procedural tasks.

%% file: section/Appendix.tex
\section{APPENDIX}
\label{appendix}
\subsection{Detailed Annotation Procedure}
\label{appendix:annotation}

As shown in Figure~\ref{fig:dataset_curation}, the annotation pipeline consists of a two-stage procedure that combines human expertise with LLM-assisted generation.
In the first stage, human annotators are required to mark time interval boundaries of fine-grained step execution status $S_s$ \revision{(i.e., \textit{just start} (at the beginning of the step), \textit{in progress} (in the middle of the step), \textit{about to finish} (near completion of the step), and \textit{step transition} (between two steps where the user transitions from one to the next))} based on the step description $D_s$ and observable hand movements.
Based on the annotated $S_s$ and $D_s$, the historical task progress $H_p$ is constructed to summarize the completed steps and task progress over time.
Annotators also label the proactive trigger $P_l$, indicating whether proactive assistance is required.
To ensure quality and consistency, a cross-validation process is adopted in which annotators review each other’s annotations.
Additionally, to avoid redundancy from visually similar frames, annotators select diverse and representative moments and filter out highly similar samples.

\begin{figure}[t]
    \centering
    \captionsetup{skip=3pt}
    \includegraphics[width=0.74\textwidth]{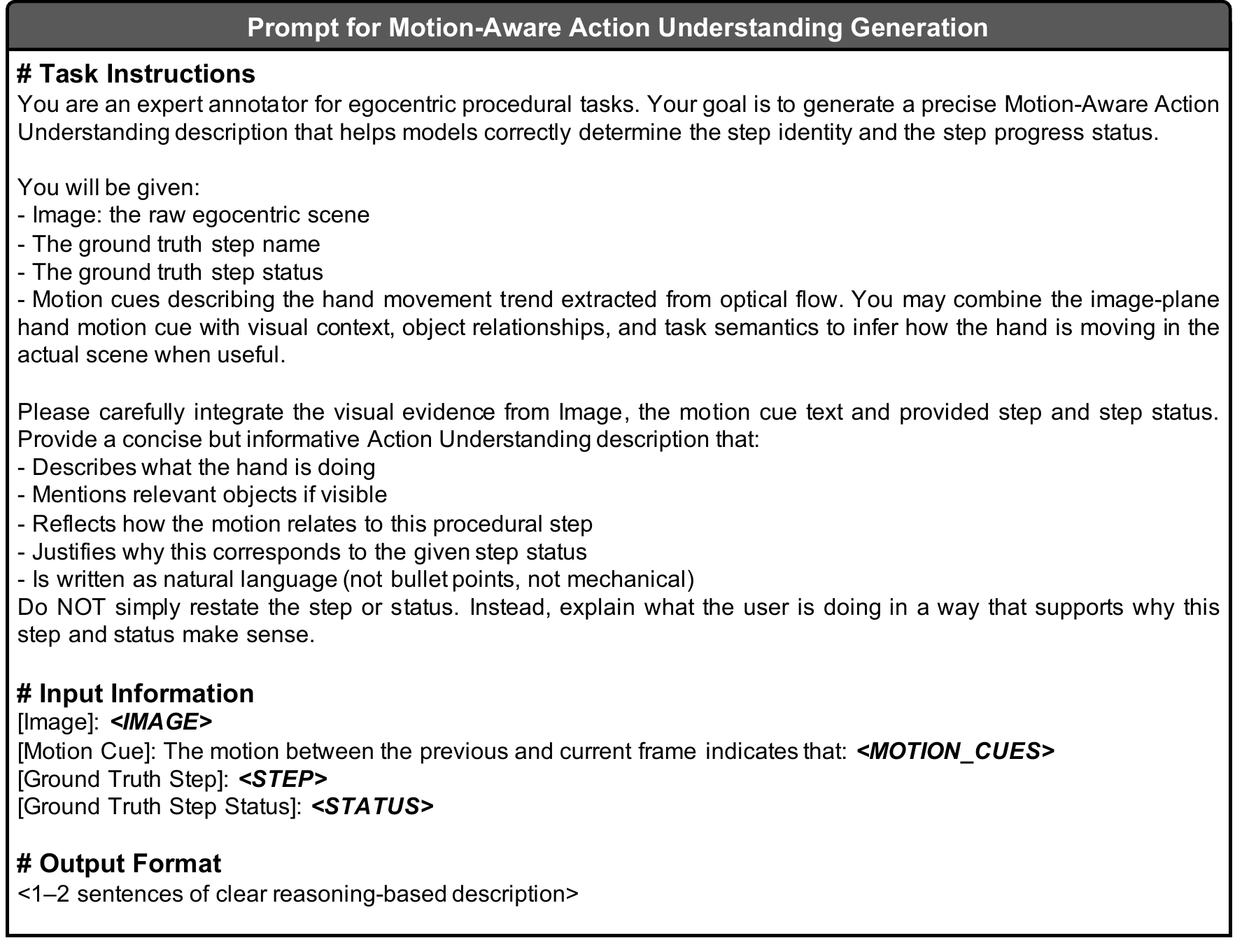}
    \caption{Prompt template for motion-aware action understanding generation.}
    \label{fig:prompt_au_generation}
\end{figure}

\begin{figure}[t]
    \centering
    \captionsetup{skip=3pt}
    \includegraphics[width=0.74\textwidth]{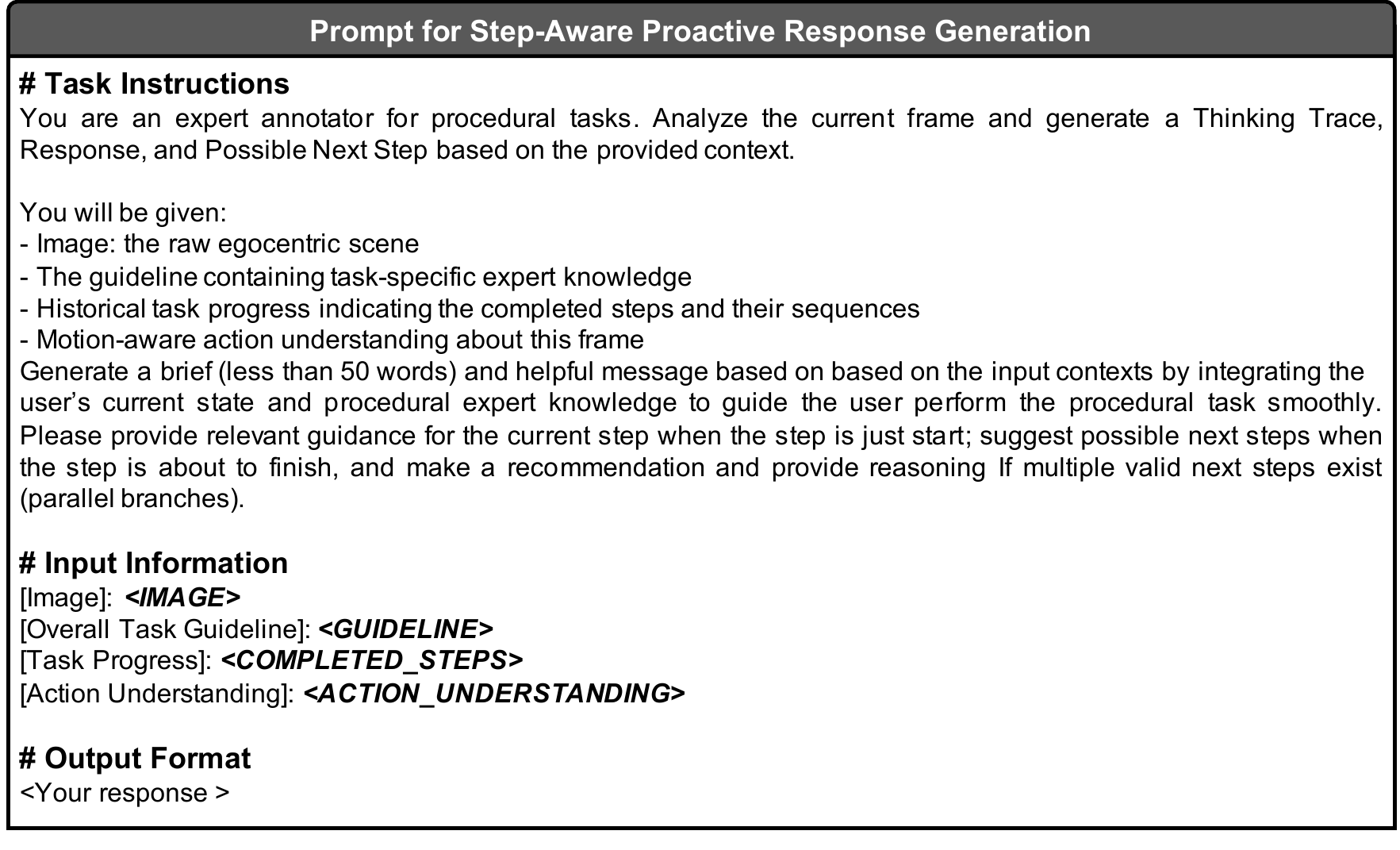}
    \caption{Prompt template for step-aware proactive response generation.}
    \label{fig:prompt_response_generation}
\end{figure}

In the second stage, advanced LLMs are employed to generate motion-aware action understanding $A_u$ and step-aware proactive responses $P_r$.
For motion-aware action understanding, the LLM is prompted with the step description $D_s$, hand motion cues $M_h$, and human-annotated step status $S_s$, together with the visual input.
The hand motion cues are extracted from each pair of consecutive frames $(I_{t-1}, I_t)$ as described in \S\ref{sec:temporal-context}.
The LLM then produces detailed action understanding annotations describing the user’s current state.
For proactive response generation, the prompt includes the historical task progress $H_p$, the generated action understanding $A_u$, and the structured procedural guideline $\mathcal{G}$ (constructed as described in \S\ref{sec:guideline}), together with the visual input.
The LLM generates contextually appropriate proactive responses $P_r$ aligned with the user’s current state.
Finally, all LLM-generated annotations $(A_u, P_r)$ are verified by two human annotators to ensure accuracy, relevance, and appropriateness.
The prompt templates used for LLM-based annotation generation are shown in Figure~\ref{fig:prompt_au_generation} and~\ref{fig:prompt_response_generation}.

\subsection{Prompts}
\label{appendix:prompts}

\begin{figure}[h]
    \centering
    \captionsetup{skip=3pt}
    \includegraphics[width=0.74\textwidth]{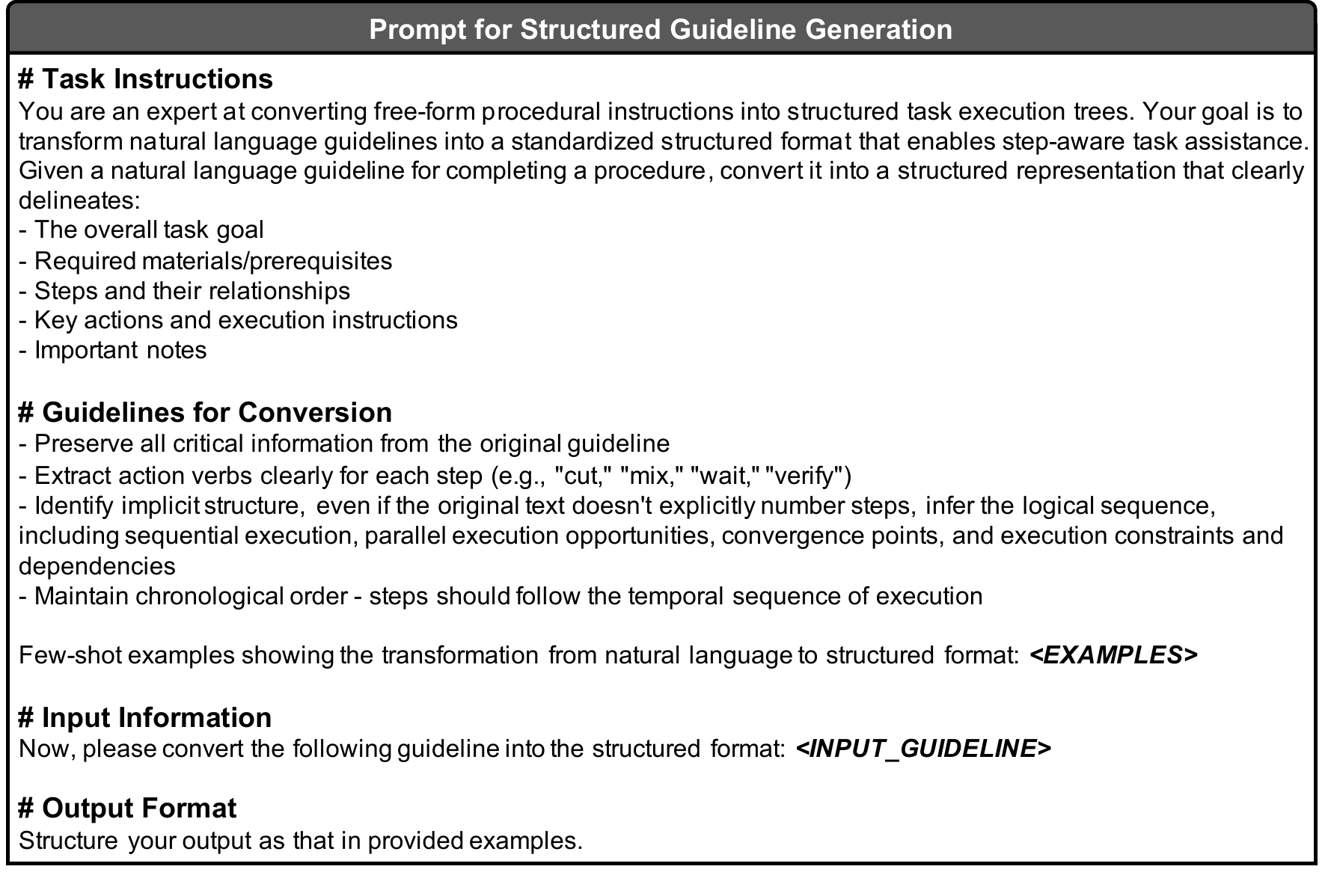}
    \caption{Prompt template for structured guideline generation.}
    \label{fig:prompt_guideline_generation}
\end{figure}

\begin{figure}[t]
    \centering
    \captionsetup{skip=3pt}
    \includegraphics[width=0.74\textwidth]{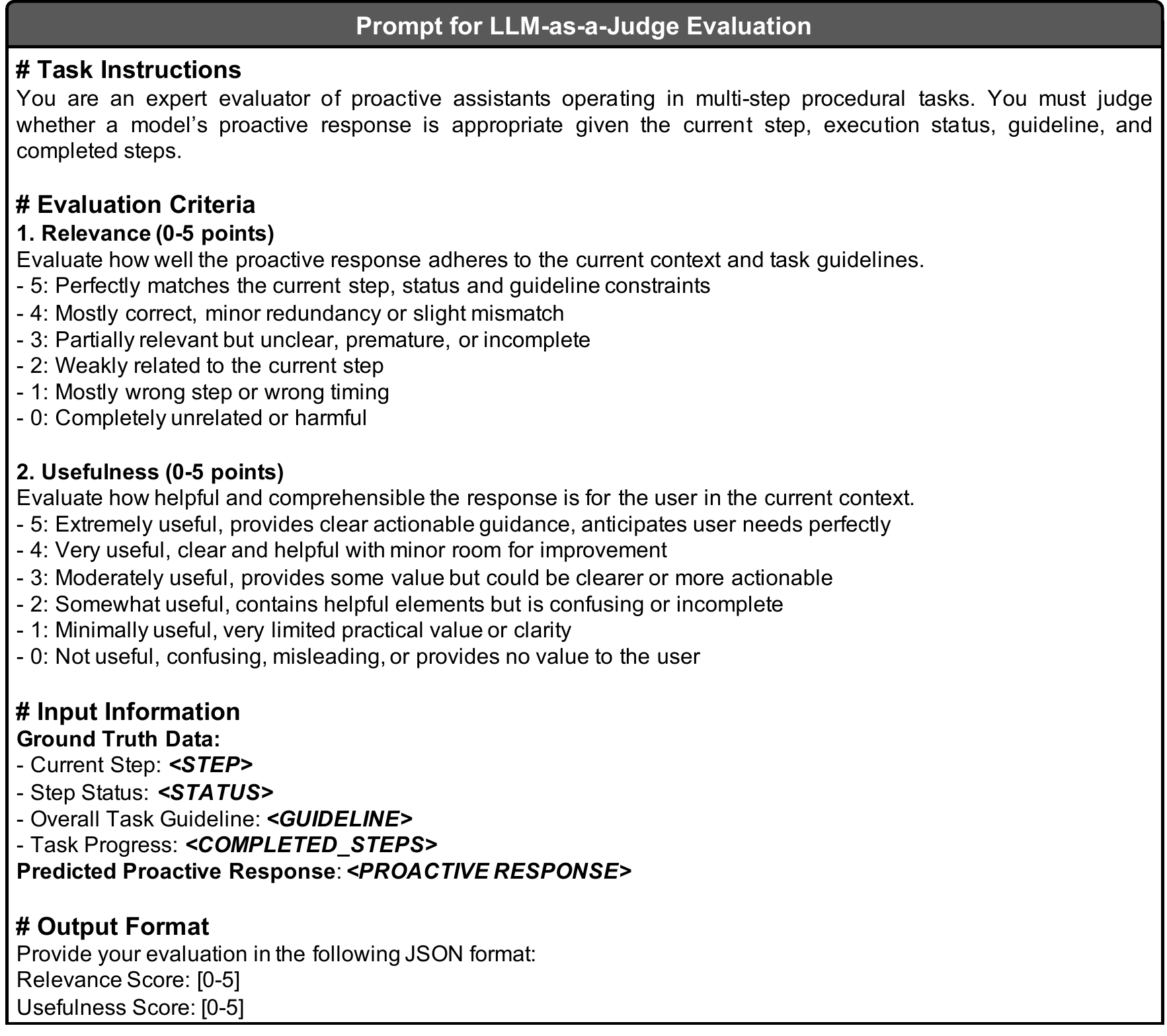}
    \caption{Prompt template for LLM-as-a-Judge Evaluation.}
    \label{fig:prompt_eval}
\end{figure}